\documentclass[journal]{IEEEtran}
\pdfoutput=1
\usepackage{cite}
\usepackage[pdftex]{graphicx}
\usepackage{amsmath}
\usepackage{amssymb}  
\usepackage{algorithmicx}
\usepackage{mathtools}
\usepackage{url}
\usepackage[table]{xcolor} 
\usepackage{threeparttable}
\usepackage[breaklinks=true,bookmarks=false]{hyperref}
\hypersetup{
    colorlinks=true,
    linkcolor=blue,
    filecolor=magenta,      
    urlcolor=cyan,
}
\usepackage{caption}
\usepackage{subcaption}
\usepackage{makecell}
\usepackage{pifont}
\usepackage{multirow}
\usepackage[linesnumbered,lined,boxed,commentsnumbered]{algorithm2e}
\usepackage{booktabs}
\usepackage{chngcntr}

\counterwithin{paragraph}{subsection}

\newcommand*{\cmark}{\ding{51}}
\newcommand*{\xmark}{\ding{55}}
\newcommand{\vc}[3]{\overset{#2}{\underset{#3}{#1}}}

\newcommand\blfootnote[1]{%
  \begingroup
  \renewcommand\thefootnote{}\footnote{#1}%
  \addtocounter{footnote}{-1}%
  \endgroup
}

\begin{document}
%
\title{No Blind Spots: Full-Surround Multi-Object Tracking for Autonomous Vehicles using \\Cameras \& LiDARs}

\author{Akshay~Rangesh$^\dagger$,~\IEEEmembership{Member,~IEEE,} and Mohan~M.~Trivedi$^\dagger$,~\IEEEmembership{Fellow,~IEEE}}

\maketitle

\blfootnote{$^\dagger$The authors are with the Laboratory for Intelligent \& Safe Automobiles, UC San Diego.}
\blfootnote{$^\ddagger$\href{https://www.youtube.com/watch?v=UowMiGXoWbc&list=PLUebh5NWCQUbP9LXdV8E_b-6y7C9hIEHT}{Video results}}
\blfootnote{$^\ddagger$\href{https://drive.google.com/open?id=1CsNyxxmHABsL0CpMtIkiF8r2S2HrKo7m}{Dataset download}}
\begin{abstract}

Online multi-object tracking (MOT) is extremely important for high-level spatial reasoning and path planning for autonomous and highly-automated vehicles. In this paper, we present a modular framework for tracking multiple objects (vehicles), capable of accepting object proposals from different sensor modalities (vision and range) and a variable number of sensors, to produce continuous object tracks. This work is a generalization of the MDP framework for MOT proposed in~\cite{xiang2015learning}, with some key extensions - First, we track objects across multiple cameras and across different sensor modalities. This is done by fusing object proposals across sensors accurately and efficiently. Second, the objects of interest (targets) are tracked directly in the \textit{real world}. This is a departure from traditional techniques where objects are simply tracked in the image plane. Doing so allows the tracks to be readily used by an autonomous agent for navigation and related tasks.

To verify the effectiveness of our approach, we test it on real world highway data collected from a heavily sensorized testbed capable of capturing full-surround information. We demonstrate that our framework is well-suited to track objects through entire maneuvers around the ego-vehicle, some of which take more than a few minutes to complete. We also leverage the modularity of our approach by comparing the effects of including/excluding different sensors, changing the total number of sensors, and the quality of object proposals on the final tracking result.


\end{abstract}

\section{Introduction}

\IEEEPARstart{T}{racking} for autonomous vehicles involves accurately identifying and localizing dynamic objects in the environment surrounding the vehicle. Tracking of surround vehicles is essential for many tasks crucial to truly autonomous driving, such as \textit{obstacle avoidance}, \textit{path planning}, and \textit{intent recognition}. To be useful for such high-level reasoning, the generated tracks should be accurate, long and robust to sensor noise. In this study, we propose a full-surround MOT framework to create such desirable tracks. 

\begin{figure}[!t]
  	\centering
	\begin{subfigure}[b]{0.4\textwidth}
		\centering
		\includegraphics[width=0.6\linewidth]{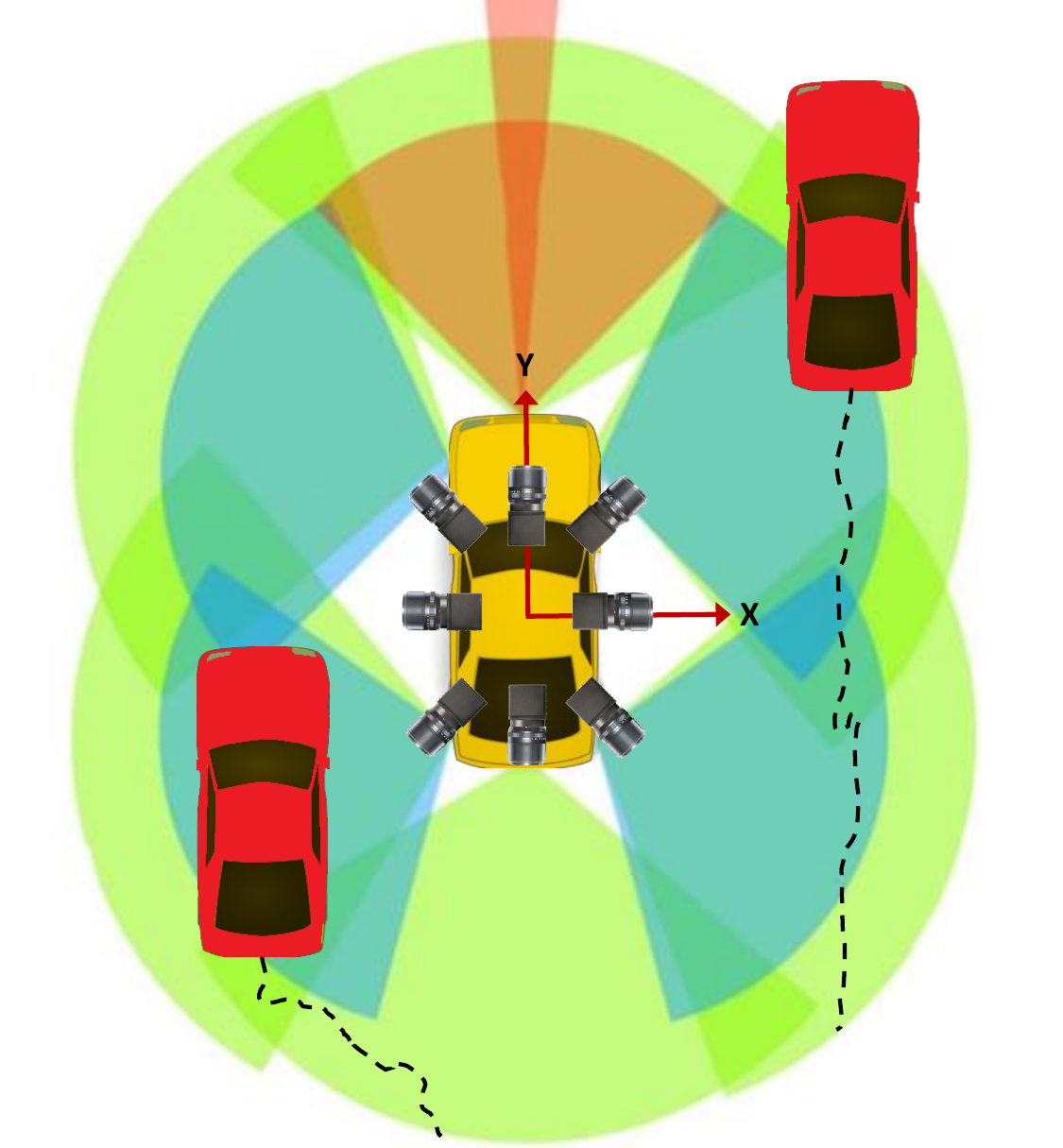}
		\caption{}
	\end{subfigure}%
\\
	\begin{subfigure}[b]{0.5\textwidth}
		\centering
		\includegraphics[width=0.6\linewidth]{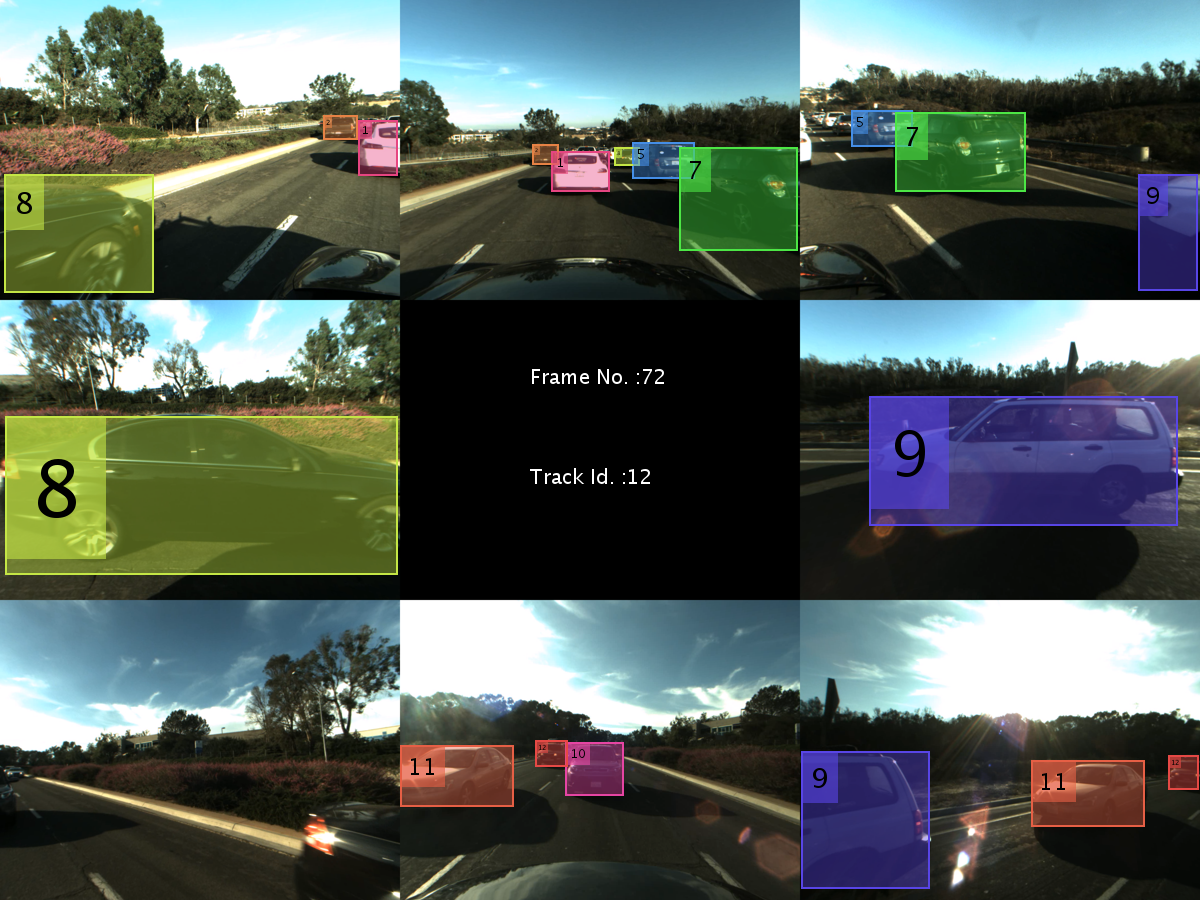}
		\caption{}
	\end{subfigure}%
\caption{\textbf{(a)} Illustration of online MOT for autonomous vehicles. The surrounding vehicles (in red) are tracked in a right-handed coordinate system centered on the ego-vehicle (center). The ego-vehicle has full-surround coverage from vision and range sensors, and must fuse proposals from each of them to generate continuous tracks (dotted lines) in the real world.
\\
\textbf{(b)} An example of images captured from a full-surround camera array mounted on our testbed, along with color coded vehicle annotations.}
\label{fig:motivation}
\end{figure}

Traditional MOT techniques for autonomous vehicles can roughly be categorized into 3 groups based on the sensory inputs they use - 1) dense point clouds from range sensors, 2) vision sensors, and 3) a fusion of range and vision sensors. Studies like \cite{choi2013multi, song2015object, asvadi2015detection, asvadi20163d1} make use of dense point clouds created by 3D LiDARs like the Velodyne HDL-64E. Such sensors, although bulky and expensive, are capable of capturing finer details of the surroundings owing to its high vertical resolution. Trackers can therefore create suitable mid-level representations like 2.5D grids, voxels etc. that retain unique statistics of the volume they enclose, and group such units together to form coherent objects that can be tracked. It must be noted however, that these approaches are reliant on having dense point representations of the scene, and would not scale well to LiDAR sensors that have much fewer scan layers. On the other hand, studies such as \cite{pfeiffer2010efficient, broggi2013full, vatavu2015stereovision, ovsep2016multi} make use of stereo vision alone to perform tracking. The pipeline usually involves estimating the disparity image and optionally creating a 3D point cloud, followed by similar mid-level representations like stixels, voxels etc. which are then tracked from frame to frame. These sensors are limited by the quality of disparity estimates and the field of view (FoV) of the stereo pair. Unlike 3D LiDAR based systems, they are unable to track objects in full-surround. There are other single camera approaches to surround vehicle behavior analysis \cite{satzoda2017vision, sivaraman2014dynamic}, but they too are limited in their FoVs and localization capabilities. Finally, there are fusion based approaches like \cite{cho2014multi, asvadi20163d, allodi2016machine}, that make use of LiDARs, stereo pairs, monocular cameras, and Radars in a variety of configurations. These techniques either perform \textit{early} or \textit{late} fusion based on their sensor setup and algorithmic needs. However, none of them seem to offer full-surround solutions for vision sensors, and are ultimately limited to fusion only in the FoV of the vision sensors.

In this study, we take a different approach to full-surround MOT and try to overcome some of the limitations in previous approaches. Specifically, we propose a framework to perform full-surround MOT using calibrated camera arrays, with varying degrees of overlapping FoVs, and with an option to include low resolution range sensors for accurate localization of objects in 3D. We term this the M$^{3}$OT framework, which stands for multi-perspective, multi-modal, multi-object tracking framework. To train and validate the M$^{3}$OT framework, we make use of naturalistic driving data collected from our testbed (illustrated in Figure~\ref{fig:motivation}) that has full-surround coverage from vision and range modalities. 

Since we use vision as our primary perception modality, we leverage recent approaches from the 2D MOT community which studies tracking of multiple objects in the 2D image plane. Recent progress in 2D MOT has focused on the \textit{tracking-by-detection} strategy, where object detections from a category detector are linked to form trajectories of the targets. To perform tracking-by-detection \textit{online} (i.e. in a causal fashion), the major challenge is to correctly associate noisy object detections in the current video frame with previously tracked objects. The basis for any data association algorithm is a similarity function between object detections and targets. To handle ambiguities in association, it is useful to combine different cues in computing the similarity, and learn an association based on these cues. Many recent 2D MOT methods such as \cite{bae2014robust, kim2012online, kuo2010multi, li2009learning, song2008vision} use some form of learning (online or offline) to accomplish data association. 
Similar to these studies, we formulate the online multi-object tracking problem using Markov Decision Processes (MDPs) proposed in \cite{xiang2015learning}, where the lifetime of an object is modeled with a MDP (see Figure~\ref{fig:MDP}), and multiple MDPs are assembled for multi-object tracking. In this method, learning a similarity function for data association is equivalent to learning a policy for the MDP. The policy learning is approached in a reinforcement learning fashion which benefits from advantages of both offline-learning and online-learning in data association. 
The M$^{3}$OT framework is also capable to naturally handle the birth/death and appearance/disappearance of targets by treating them as state transitions in the MDP, and also benefits from the strengths of online learning approaches in single object tracking (\cite{babenko2011robust, bao2012real, hare2016struck, kalal2012tracking}). 

\begin{figure*}[ht!]
\begin{center}
\includegraphics[width=0.9\linewidth]{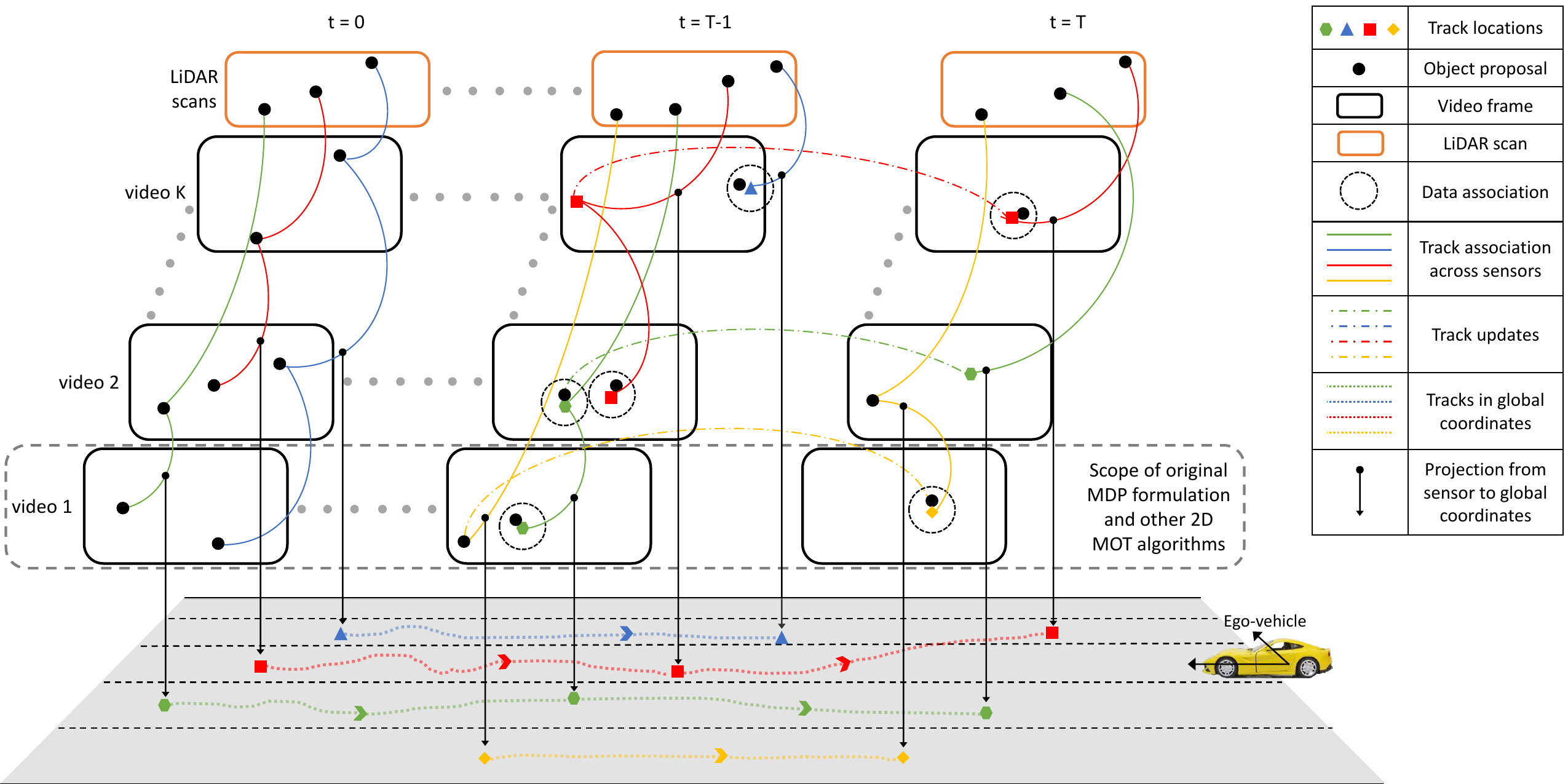}
\end{center}
\caption{Illustration of proposed M$^3$OT framework and its scope in comparison to traditional 2D MOT algorithms. Data is associated not only within video frames, but also across other videos and sensors. The algorithm produces tracks in each individual sensor coordinate frame, and in the desired global coordinate frame using cross-modal fusion in an online manner.}
\label{fig:block_diag}
\end{figure*}

Our main contributions in this work can be summarized as follows - 1) We extend and improve the MDP formulation originally proposed for 2D MOT~\cite{xiang2015learning}, and modify it to track objects in 3D (real world). 2) We make the M$^{3}$OT framework capable of tracking objects across multiple vision sensors in calibrated camera arrays by carrying out efficient and accurate fusion of object proposals. 3) The M$^{3}$OT framework is made highly \textit{modular}, capable of working with any number of cameras, with varying degrees of overlapping FoVs, and with the option to include range sensors for improved localization and fusion in 3D. The above contributions and the wider scope and applicability of this work in comparison to traditional 2D MOT approaches are highlighted in Figure~\ref{fig:block_diag}. Finally, we carry out experiments using naturalistic driving data collected on highways using full-surround sensory modalities, and validate the accuracy, robustness and modularity of our framework.

\begin{table*}[ht!]
\centering
\resizebox{0.9\linewidth}{!}{%
\begin{threeparttable}
\caption{Relevant research in 3D MOT for intelligent vehicles.}
\begin{tabular}{| c | c | c | c | c | c | c | c | c | c | p{4cm} |}
\hline
\multirow{3}{*}{\textbf{Study}} & \multicolumn{5}{c |}{\textbf{Sensors used}} & \multicolumn{3}{c |}{\textbf{Tracker Type}} & \multicolumn{2}{c |}{\textbf{Experimental Analysis}}\\ 
\cline{2-11}
 & \thead{Monocular\\camera} & \thead{Stereo\\pair} & \thead{Full-surround\\camera array} & \thead{LiDAR\\ } & \thead{Radar\\ } & \thead{Single object\\tracker} & \thead{Multi-object\\tracker} & \thead{Online\\(causal)} & \thead{Dataset\\ } & \thead{Evaluation metrics}\\
\hline \hline 
Choi et al.\cite{choi2013multi} & - & - & - & \cmark & - & - & \cmark & \cmark & Proposed & \makecell{Distance and velocity errors}\\ \hline
Broggi et al.\cite{broggi2013full} & - & \cmark & - & - & - & - & \cmark & \cmark & Proposed & \makecell{True positives, false positives,\\false negatives}\\ \hline
Song et al.\cite{song2015object} & - & - & - & \cmark & - & \cmark & \xmark & \cmark & KITTI & \makecell{Position error, intersection ratio}\\ \hline 
Cho et al.\cite{cho2014multi} & \cmark & - & - & \cmark & \cmark & - & \cmark & \cmark & Proposed & \makecell{Correctly tracked, falsely tracked,\\true and false positive rates}\\ \hline  
Asvadi et al.\cite{asvadi2015detection} & - & - & - & \cmark & - & - & \cmark & \cmark & KITTI & - \\ \hline 
Vatavu et al.\cite{vatavu2015stereovision} & - & \cmark & - & - & - & - & \cmark & \cmark & Proposed & - \\ \hline 
Asvadi et al.\cite{asvadi20163d1} & - & - & - & \cmark & - & - & \cmark & \cmark & KITTI & \makecell{Number of missed and false\\obstacles}\\ \hline 
Asvadi et al.\cite{asvadi20163d} & \cmark & - & - & \cmark & - & \cmark & \xmark & \cmark & KITTI & \makecell{Average center location errors in\\2D and 3D, orientation errors}\\ \hline
O{\v{s}}ep et al.\cite{ovsep2016multi} & - & \cmark & - & - & - & - & \cmark & \cmark & KITTI & \makecell{Class accuracy, GOP recall,\\tracking recall}\\ \hline
Allodi et al.\cite{allodi2016machine} & - & \cmark & - & \cmark & - & - & \cmark & \cmark & KITTI & \makecell{MOTP, IDS, Frag, average\\localization error}\\ \hline
Dueholm et al.\cite{dueholm2016trajectories} & - & - & \cmark & - & - & - & \cmark & \xmark & \makecell{VIVA Surround} & \makecell{MOTA, MOTP, IDS, Frag,\\Precision, Recall}\\ \hline
\textbf{\makecell{This work\tnote{1}\\(M$^{3}$OT)}} & - & - & \cmark & \cmark & - & - & \cmark & \cmark & Proposed & \makecell{MOTA, MOTP, MT, ML, IDS for a\\variety of sensor configurations}\\ \hline
\end{tabular}
\begin{tablenotes}
\item[1] this framework can work with or without LiDAR sensors, and with any subset of camera sensors.
\end{tablenotes}
\label{tab:related_work}
\end{threeparttable}
}
\end{table*}

\section{Related Research}\label{related_work}

We highlight some representative works in 2D and 3D MOT below. We also summarize the some key aspects of related 3D MOT studies in Table~\ref{tab:related_work}. For a recent survey on detection and tracking for intelligent vehicles, we refer the reader to \cite{sivaraman2013looking}.

\textbf{2D MOT:} Recent research in MOT has focused on tracking-by-detection, where the main challenge is data association for linking object detections to targets. Majority of batch (\textit{offline}) methods (\cite{berclaz2011multiple, butt2013multi, li2009learning, milan2014continuous, niebles2010efficient, pirsiavash2011globally, zhang2008global}) formulate MOT as a global optimization problem in a graph-based representation, while \textit{online} methods solve the data association problem either probabilistically \cite{khan2005mcmc, oh2009markov, okuma2004boosted} or deterministically (e.g., Hungarian algorithm \cite{munkres1957algorithms} in \cite{bae2014robust, kim2012online} or greedy association \cite{breitenstein2011online}). A core component in any data association algorithm is a similarity function between objects. Both batch methods \cite{kuo2010multi, li2009learning} and online methods \cite{bae2014robust, kim2012online, song2008vision} have explored the idea of learning to track, where the goal is to learn a similarity function for data association from training data. In this work, we extend and improve the MDP framework for 2D MOT proposed in \cite{xiang2015learning}, which is an online method that uses reinforcement learning to learn associations.

\textbf{3D MOT for autonomous vehicles:} In \cite{pfeiffer2010efficient}, the authors use a stereo rig to first calculate the disparity using Semi-Global Matching (SGM), followed by height based segmentation and free-space calculation to create a mid-level representation using \textit{stixels} that encode the height within a cell. Each stixel is then represented by a 6D state vector, which is tracked using the Extended Kalman Filter (EKF). In \cite{broggi2013full}, the authors use a \textit{voxel} based representation instead, and cluster neighboring voxels based on color to create objects that are then tracked using a greedy association model. The authors in \cite{vatavu2015stereovision} use a grid based representation of the scene, where cells are grouped to create objects, each of which is represented by a set of control points on the object surface. This creates a high dimensional state-space representation, which is accounted for by a Rao-Blackwellized particle filter. More recently, the authors of \cite{ovsep2016multi} propose to carry out semantic segmentation on the disparity image, which is then used to generate generic object proposals by creating a scale-space representation of the density, followed by multi-scale clustering. The proposed clusters are then tracked using a Quadratic Pseudo-Boolean Optimization (QPBO) framework. The work in \cite{dueholm2016trajectories} use a camera setup similar to ours, but the authors propose an offline framework for tracking, hence limiting their use to surveillance related applications.

Alternatively, there are approaches that make use of dense point clouds generated by LiDARs as opposed to creating point clouds from a disparity image. In \cite{choi2013multi}, the authors first carry out ground classification based on the variance in the radius of each scan layer, followed by a 2.5D occupancy grid representation of the scene. The grid is then segmented, and regions of interest (RoIs) identified within, each of which is tracked by a standard Kalman Filter (KF). Data association is achieved by simple global nearest neighbor. Similar to this, the authors in \cite{asvadi2015detection} use a 2.5D occupancy grid based representation, but augment this with an occupancy grid \textit{map} which accumulates past grids to create a coherent global map of occupancy by accounting for ego-motion. Using these two representations, a 2.5D \textit{motion} grid is created by comparing the map with the latest occupancy grid, which isolates and identifies dynamic objects in the scene. Although the work in \cite{asvadi20163d1} follows the same general idea, the authors propose a piece-wise ground plane estimation scheme capable of handling non-planar surfaces. In a departure from grid based methods, the authors in \cite{song2015object} project the 3D point cloud onto a virtual image plane, creating an object appearance model based on 4 image-based cues for each template of the desired target. A particle filtering framework is implemented, where the particle with least reconstruction error with respect to the stored template is chosen to update the tracker. Background filtering and occlusion detection are implemented to improve performance.

Finally, we list recent methods that rely on fusion of different sensor modalities to function. In \cite{cho2014multi}, the authors propose an EKF based fusion scheme, where measurements from each modality are fed sequentially. Vision is used to classify the object category, which is then used to choose appropriate motion and observation models for the object. Once again, observations are associated based on a global nearest neighbor policy. This is somewhat similar to the work in \cite{asvadi20163d}, where given an initial 3D detection box, the authors propose to project 3D points within the box to the image plane and calculate the convex hull of projected points. In this case, a KF is used to perform both fusion and tracking, where fusion is carried out by projecting the 2D hull to a sparse 3D cloud, and using both 3D cues to perform the update. In contrast, the authors of \cite{allodi2016machine} propose using the Hungarian algorithm (for bipartite matching) for both data association and fusion of object proposals from different sensors. The scores for association are obtained from Adaboost classifiers trained on high-level features. The objects are then tracked using an Unscented Kalman Filter (UKF).

\section{Fusion of Object Proposals}\label{fusion}

In this study, we make use of full-surround camera arrays comprising of sensors with varying FoVs. The M$^{3}$OT framework, however, is capable of working with any type and number of cameras, as long as they are calibrated. In addition to this, we also propose a variant of the framework for cases where LiDAR point clouds are available. To effectively utilize all available sensors, we propose an \textit{early fusion} of object proposals obtained from each of them. At the very start of each time step during tracking, we identify and fuse all proposals belonging to the same object. These proposals are then utilized by the M$^{3}$OT framework to carry out tracking. It must be noted that this usage of ``early fusion" is in contrast to the traditional usage of the term to refer to fusion of raw sensor data to provide a merged representation. 

\paragraph{Projection \& Back-projection}
It is essential to have a way of associating measurements from different sensors to track objects across different camera views, and to carry out efficient fusion across sensor modalities. This is achieved by defining a set of \textit{projection} mappings, one from each sensor's unique coordinate system to the global coordinate system, and a set of \textit{back-projection} mappings that take measurements in the global coordinate system to individual coordinate systems. In our case, the global coordinate system is centered at the mid-point of the rear axle of the ego-vehicle. The axes form a right-handed coordinate system as shown in Figure~\ref{fig:motivation}.

The LiDAR sensors output a 3D point cloud in a common coordinate system at every instant. This coordinate frame may either be centered about a single LiDAR sensor, or elsewhere depending on the configuration. In this case, the projection and back-projection mappings are simple 3D coordinate transformations:
\begin{equation}
P_{range \rightarrow G}(\mathbf{x}^{range}) = \boldsymbol{R}_{range} \cdot \mathbf{x}^{range} + \mathbf{t}_{range} \mathbf{l},
\end{equation}
and,
\begin{equation}
P_{range \leftarrow G}(\mathbf{x}^{G}) = \boldsymbol{R}_{range}^T \cdot \mathbf{x}^{G} -\boldsymbol{R}_{range}^T \mathbf{t}_{range},
\end{equation}
where $P_{range \rightarrow G}(\cdot)$ and $P_{range \leftarrow G}(\cdot)$ are the projection and back-projection mappings from the LiDAR (range) coordinate system to the global (G) coordinate system and vice-versa. The vectors $\mathbf{x}^{range}$ and $\mathbf{x}^{G}$ are the corresponding coordinates in the LiDAR and global coordinate frames. The $3 \times 3$ orthonormal rotation matrix $\boldsymbol{R}_{range}$ and translation vector $\mathbf{t}_{range}$ are obtained through calibration.

Similarly, the back-projection mappings for each camera $k \in \{1, \cdots, K\}$ can be defined as:
\begin{align}
P_{cam_k \leftarrow G}(\mathbf{x}^{G}) = (u, v)^T,\\
\text{s.t }\begin{bmatrix}u \\v \\1 \end{bmatrix} = \boldsymbol{C}_k \cdot [\boldsymbol{R}_k | \mathbf{t}_k] \cdot [(\mathbf{x}^G)^T | 1]^T,
\end{align}
where the set of camera calibration matrices $\{\boldsymbol{C}_k\}_{k=1}^K$ are obtained after the intrinsic calibration of cameras, the set of tuples $\{(\boldsymbol{R}_k, \mathbf{t}_k)\}_{k=1}^K$ obtained after extrinsic calibration, and $(u, v)^T$ denotes the pixel coordinates after back-projection.

Unlike the back-projection mappings, the projection mappings for camera sensors are not well defined. In fact, the mappings are one-to-many due to the depth ambiguity of single camera images. To find a good estimate of the projection, we use two different approaches. In case of a vision-only system, we use the \textit{inverse perspective mapping} (IPM) approach:
\begin{align}
P_{cam_k \rightarrow G}(\mathbf{x}^{k}) = (x, y)^T,\\
\text{s.t. }\begin{bmatrix}x \\y \\1 \end{bmatrix} = \boldsymbol{H}_k \cdot [(\mathbf{x}^k)^T | 1]^T,
\end{align}
where $\{\boldsymbol{H}_k\}_{k=1}^K$ are the set of homographies obtained after IPM calibration. Since we are only concerned with lateral and longitudinal displacements of vehicles in the global coordinate system, we only require the $(x, y)^T$ coordinates, and set the altitude coordinate to a fixed number. 

\begin{figure}[!t]
  	\centering
	\begin{subfigure}[b]{0.5\textwidth}
		\centering
		\includegraphics[width=0.7\linewidth]{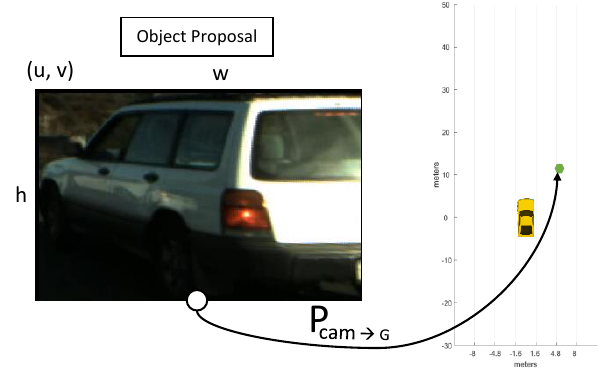}
		\caption{}
		\label{fig:ipm_projection}
	\end{subfigure}%
\\
	\begin{subfigure}[b]{0.5\textwidth}
		\centering
		\includegraphics[width=0.75\linewidth]{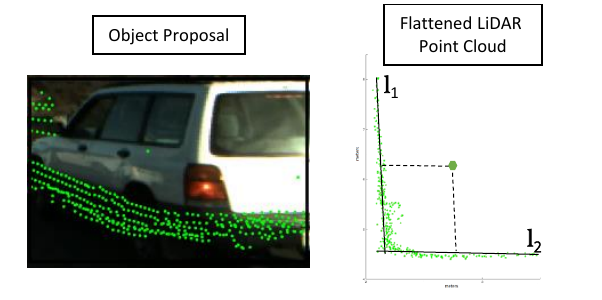}
		\caption{}
		\label{fig:lidar_projection}
	\end{subfigure}%
\caption{Projection of object proposals using: (a) \textbf{IPM:} The bottom center of the bounding boxes are projected into the global coordinate frame (right), (b) \textbf{LiDAR point clouds:} LiDAR points that fall within a detection window are flattened and lines are fit to identify the vehicle center (right).}
\end{figure}

\paragraph{Sensor Measurements \& Object Proposals}
As we adopt a tracking-by-detection approach, each sensor is used to produce object proposals to track and associate. In case of vision sensors, a vehicle detector is run on each individual camera's image to obtain multiple detections $d$, each of which is defined by a bounding box in the corresponding image. Let $(u, v)$ denote the top left corner and $(w, h)$ denote the width and height of a detection $d$ respectively. 

In case of a vision-only system, the corresponding location of $d$ in the global coordinate system is obtained using the mapping $P_{cam_k \rightarrow G}((u+\frac{w}{2}, v+h)^T)$, where $k$ denotes the camera from which the proposal was generated. This procedure is illustrated in Figure~\ref{fig:ipm_projection}. Alternatively, this could be replaced by a purely vision based 3D detector like the one proposed in~\cite{rangesh2018ground}, where both the global pose and 2D bounding box of an object are obtained from the same algorithm.

In cases where LiDAR sensors are available, an alternative is considered (shown in Figure~\ref{fig:lidar_projection}). First, the back-projected LiDAR points that fall within a detection box $d$ are identified using a look-up table with pixel coordinates as the key, and the corresponding global coordinates as the value. These points are then flattened by ignoring the altitude component of their global coordinates. Next, a line $\mathbf{l}_1$ is fitted to these points using RANSAC with a small inlier ratio (0.3). This line aligns with the dominant side of the detected vehicle. The other side of the vehicle corresponding to line $\mathbf{l}_2$ is then identified by removing all inliers from the previous step and repeating a RANSAC line fit with a slightly higher inlier ratio (0.4). Finally, the intersection of lines $\mathbf{l}_1$ and $\mathbf{l}_2$ along with the vehicle dimensions yield the global coordinates of the vehicle center. The vehicle dimensions are calculated based on the extent of the LiDAR points along a given side of the vehicle, and stored for each track separately for later use.

Depending on the type of LiDAR sensors used, object proposals along with their dimensions in the real world can be obtained. However, we decide not to make use of LiDAR proposals, but rather use vision-only proposals with high recall by trading off some of the precision. This was seen to provide sufficient proposals to track all surrounding vehicles, at the expense of more false positives which the tracker is capable of handling.

\paragraph{Early Fusion of Proposals}
Since we operate with camera arrays with overlapping FoVs, the same vehicle may be detected in two adjacent views. It is important to identify and \textit{fuse} such proposals to track objects across camera views. Once again, we propose two different approaches to carry out this fusion. For vision-only systems, the fusion of proposals is carried out in 4 steps: i) Project proposals from all cameras to the global coordinate system using proposed mappings, ii) Sort all proposals in descending order based on their confidence scores (obtained from the vehicle detector), iii) Starting with the highest scoring proposal, find the subset of proposals whose euclidean distance in the global coordinate system falls within a predefined threshold. These proposals are considered to belong to the same object and removed from the original set of proposals. In practice, we use a threshold of $1m$ for grouping proposals. iv) The projection of each proposal within this subset is set to the mean of projections of all proposals within the subset. This process is repeated for the remaining proposals until no proposals remain.

\begin{figure}[!t]
  	\centering
	\includegraphics[width=0.9\linewidth]{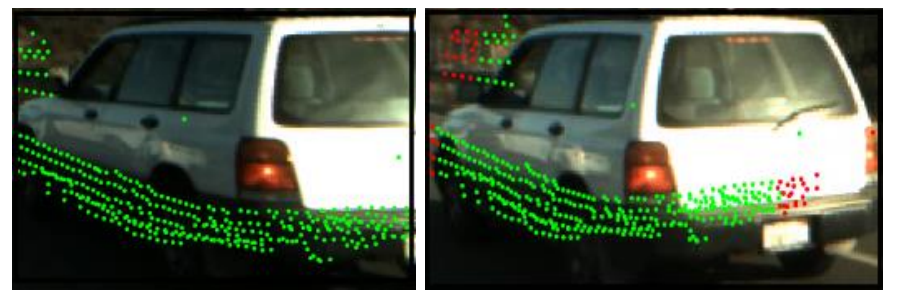}
\caption{Fusion of object proposals using LiDAR point clouds: Points common to both detections are drawn in green, and the rest are drawn red.}
\label{fig:fusion_lidar}
\end{figure}

Alternatively, for a system consisting of LiDAR sensors, we project the 3D point cloud onto each individual camera image. Next, for each pair of proposals, we make a decision as to whether or not they belong to the same object. This is done by considering the back-projected LiDAR points that fall within the bounding box of each proposal (see Figure~\ref{fig:fusion_lidar}). Let $\mathcal{P}_1$ and $\mathcal{P}_2$ denote the index set of LiDAR points falling within each bounding box. Then, two proposals are said to belong to the same object if:
\begin{equation}
\text{max}\bigg(\frac{\mathcal{P}_1 \cap \mathcal{P}_2}{|\mathcal{P}_1|}, \frac{\mathcal{P}_1 \cap \mathcal{P}_2}{|\mathcal{P}_2|}\bigg) \geq 0.8,
\end{equation}

where $|\mathcal{P}_1|$ and $|\mathcal{P}_2|$ denote the cardinalities of sets $\mathcal{P}_1$ and $\mathcal{P}_2$ respectively. It should be noted that after fusion is completed, the union of LiDAR point sets that are back-projected into fused proposals can be used to obtain better projections.

\section{M$^{3}$OT Framework}\label{framework}

\begin{figure}[t]
\begin{center}
\includegraphics[width=0.75\linewidth]{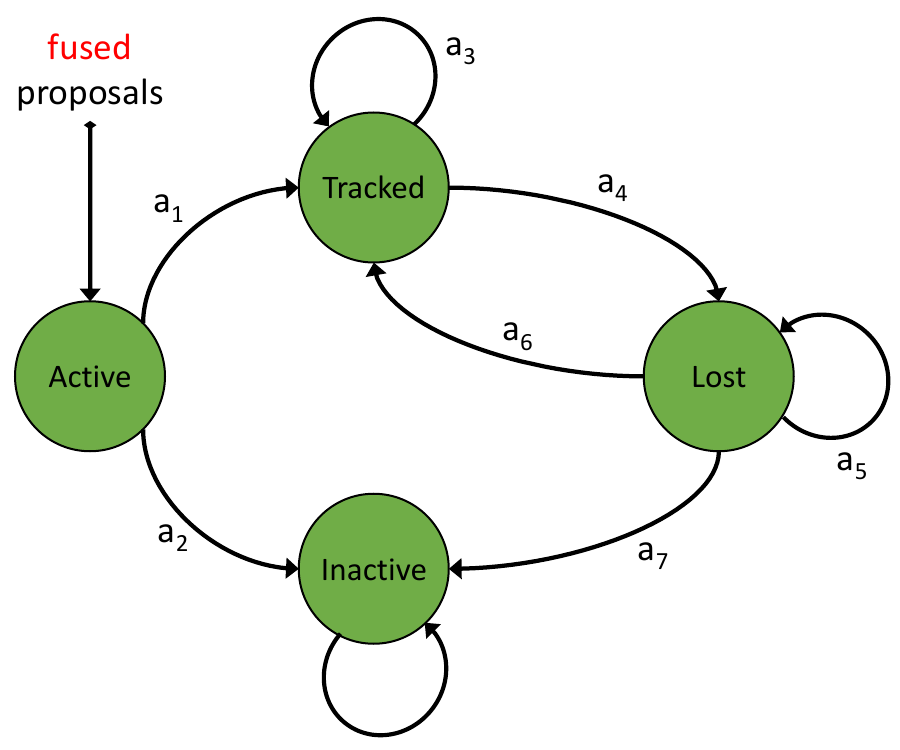}
\end{center}
\caption{The Markov Decision Process (MDP) framework proposed in \cite{xiang2015learning}. In this work, we retain the structure of the MDP, and modify the actions, rewards and inputs to enable multi-sensory tracking.}
\label{fig:MDP}
\end{figure}

Once we have a set of fused object proposals, we feed them into the MDP as illustrated in Figure~\ref{fig:MDP}. Although the MDP framework introduced in~\cite{xiang2015learning} forms a crucial building block of the proposed M$^3$OT framework, we believe that extending this to account for multiple sensors and modalities is a non-trivial endeavor (see Figure~\ref{fig:block_diag}). This section highlights and explains the modifications we propose to achieve these objectives.

\subsection{Markov Decision Process} \label{MDP}
As detailed in \cite{xiang2015learning}, we model the lifetime of a target with a Markov Decision Process (MDP). The MDP consists of the tuple $(\mathcal{S}, \mathcal{A}, T(\cdot, \cdot), R(\cdot, \cdot))$, where:
\begin{itemize}
\item States $s \in \mathcal{S}$ encode the status of the target.
\item Actions $a \in \mathcal{A}$ define the actions that can be taken.
\item The state transition function $T: \mathcal{S} \times \mathcal{A} \mapsto \mathcal{S}$ dictates how the target transitions from one state to another, given an action.
\item The real-valued reward function $R: \mathcal{S} \times \mathcal{A} \mapsto \mathbb{R}$ assigns the immediate reward received after executing action $a$ in state $s$.
\end{itemize}

In this study, we retain the states $\mathcal{S}$, actions $\mathcal{A}$ and the state transition function $T(\cdot, \cdot)$ from the MDP framework for 2D MOT \cite{xiang2015learning}, while changing only the reward function $R(\cdot, \cdot)$.

\textbf{States:} The state space is partitioned into four subspaces, i.e., $\mathcal{S} = \mathcal{S}_{Active} \cup \mathcal{S}_{Tracked} \cup \mathcal{S}_{Lost} \cup \mathcal{S}_{Inactive}$, where each subspace contains infinite number of states which encode the information of the target depending on the feature representation, such as appearance, location, size and history of the target. Figure~\ref{fig:MDP} illustrates the transitions between the four subspaces. \textit{Active} is the initial state for any target. Whenever an object is detected by the object detector, it enters an \textit{Active} state. An active target can transition to \textit{Tracked} or \textit{Inactive}. Ideally, a true positive from the object detector should transition to a \textit{Tracked} state, while a false alarm should enter an \textit{Inactive} state. A tracked target can stay tracked, or transition to a \textit{Lost} state if the target is not visible due to some reason, e.g. occlusion, or disappearance from sensor range. Likewise, a lost target can stay \textit{Lost}, or go back to a \textit{Tracked} state if it appears again, or transition to an \textit{Inactive} state if it has been lost for a sufficiently long time. Finally, \textit{Inactive} is the terminal state for any target, i.e., an inactive target stays inactive forever.

\textbf{Actions and Transition Function:} Seven possible transitions are designed between the state subspaces, which correspond to seven actions in our target MDP. Figure~\ref{fig:MDP} illustrates these transitions and actions. In the MDP, all the actions are deterministic, i.e., given the current state and an action, we specify a new state for the target. For example, executing action $a_6$ on a \textit{Lost} target would transfer the target into a \textit{Tracked} state, i.e., $T(s_{Lost}, a_6) = s_{Tracked}$. 

\textbf{Reward Function:} As in the original study \cite{xiang2015learning}, we learn the reward function from training data, i.e., an inverse reinforcement learning problem, where we use ground truth trajectories of the targets as supervision.

\subsection{Policy}

\paragraph{Policy in Active States}
In an Active state $s$, the MDP makes the decision between transferring an object proposal into a Tracked or Inactive state based on whether the detection is true or noisy. To do this, we train a set of binary Support Vector Machines (SVM) offline, one for each camera view, to classify a detection belonging to that view into Tracked or Inactive states using a normalized 5D feature vector $\phi_{Active}(s)$, i.e., 2D image plane coordinates, width, height and score of the detection, where training examples are collected from training video sequences.

This is equivalent to learning the reward function:
\begin{equation}
R_{Active}(s, a) = y(a)((\mathbf{w}_{Active}^k)^T \cdot \phi_{Active}(s) + b_{Active}^k),
\end{equation}
for an object proposal belonging to camera $k \in \{1, \cdots, K\}$. $(\mathbf{w}_{Active}^k, b_{Active}^k)$ defines the learned weights and bias of the SVM for camera $k$, $y(a) = +1$ if action $a = a_1$, and $y(a) = -1$ if $a = a_2$ (see Figure~\ref{fig:MDP}). Training a separate SVM for each camera view allows weights to be learned based on object dimensions and locations in that particular view, and thus works better than training a single SVM for all views. Since a single object can result in multiple proposals, we initialize a tracker for that object if any of the fused proposals result in a positive reward. Note that a false positive from the object detector can still be misclassified and transferred to a Tracked state, which we then leave to be handled by the MDP in Tracked and Lost states.

\paragraph{Policy in Tracked States}
In a \textit{Tracked} state, the MDP needs to decide whether to keep tracking the target or to transfer it to a \textit{Lost} state. As long as the target is visible, we should keep tracking it. Else, it should be marked ``lost". We build an appearance model for the target online and use it to track the target. If the appearance model is able to successfully track the target in the next video frame, the MDP leaves the target in a Tracked state. Otherwise, the target is transferred to a Lost state. 

\textbf{Template Representation:} The appearance of the target is simply represented by a template that is an image patch of the target in a video frame. Whenever an object detection is transferred to a Tracked state, we initialize the target template with the detection bounding box. If the target is initialized with multiple fused proposals, then each detection is stored as a template. We make note of detections obtained from different camera views, and use these to model the appearance of the target in that view. This is crucial to track objects across camera views under varying perspective changes. When the target is being tracked, the MDP collects its templates in the tracked frames to represent the history of the target, which will be used in the Lost state for decision making.

\textbf{Template Tracking:} 
Tracking of templates is carried out by performing dense optical flow as described in \cite{xiang2015learning}. The stability of the tracking is measured using the median of the Forward-Backward (FB) errors~\cite{kalal2012tracking} of all sampled points: $e_{medFB} = median({e(\mathbf{u}_i)}_{i=1}^n)$, where $\mathbf{u}_i$ denotes each sampled point, and $n$ is the total number of points. If $e_{medFB}$ is larger than some threshold, the tracking is considered to be unstable. Moreover, after filtering out unstable matches whose FB error is larger than the threshold, a new bounding box of the target is predicted using the remaining matches by measuring scale change between points before and after. This process is carried out for all camera views in which a target template has been initialized and tracking is in progress.

Similar to the original MDP framework, we use the optical flow information in addition to the object proposals history to prevent drifting of the tracker. To do this, we compute the bounding box overlap between the target box for $l$ past frames, and the corresponding detections in each of those frames. Then we compute the mean bounding box overlap for the past $L$ tracked frames $o_{mean}$ as another cue to make the decision. Once again, this process is repeated for each camera view the target is being tracked in. In addition to the above features, we also \textit{gate} the target track. This involves introducing a check to see if the current global position of the tracked target falls within a window (gate) of it's last known global position. This forbids the target track from latching onto objects that appear close on the image plane, yet are much farther away in the global coordinate frame. We denote the last know global position and the currently tracked global position of the target as $\mathbf{x}^{G}(t-1)$ and $\mathbf{\hat{x}}^{G}(t)$ respectively.

Finally, we define the reward function in a Tracked state $s$ using the feature set $\phi_{Tracked}(s) = (\{e_{medFB}^{k'}\}_{k'=1}^{K'}, \{o_{mean}^{k'}\}_{k'=1}^{K'}, \mathbf{x}^{G}(t-1), \mathbf{\hat{x}}^{G}(t))$ as:
\begin{equation}
R_{Tracked}(s, a) = \begin{cases}
					y(a), & \text{if }\exists k' \in \{1, \cdots, K'\}\text{ s.t.}\\
					 & (e_{medFB}^{k'} < e_0) \land (o_{mean}^{k'} > o_0)\\
					 & \land (|\mathbf{x}^{G}(t-1) - \mathbf{\hat{x}}^{G}(t)| \leq \mathbf{t}_{gate}),\\
					-y(a), & \text{otherwise},
					\end{cases}
\end{equation}
where $e_0$ and $o_0$ are fixed thresholds, $y(a) = +1$ if action $a = a_3$, and $y(a) = -1$ if $a = a_4$ (see Figure~\ref{fig:MDP}). $k'$ above indexes camera views in which the target is currently being tracked and $\mathbf{t}_{gate}$ denotes the gating threshold. So the MDP keeps the target in a Tracked state if $e_{medFB}$ is smaller and $o_{mean}$ is larger than their respective thresholds for any one of $K'$ camera views in addition to satisfying the gating check. Otherwise, the target is transferred to a Lost state. 

\textbf{Template Updating:} The appearance model of the target needs to be regularly updated in order to accommodate appearance changes. As in the original work, we adopt a``lazy" updating rule and resort to the object detector in preventing tracking drift. 
This is done so that we don't accumulate tracking errors, but rather rely on data association to handle appearance changes and continue tracking. In addition to this, templates are initialized in views where the target is yet to be tracked by using proposals that are fused with detections corresponding to the tracked location in an adjacent camera view. This helps track objects that move across adjacent camera views, by creating target templates in the new view as soon as they are made available.

\IncMargin{1em}
\begin{algorithm}[t!]
\smaller
\SetKwInOut{Input}{input}\SetKwInOut{Output}{output}
\Input{Set of multi-video sequences $\mathcal{V} = \{(v_i^1, \cdots, v_i^K)\}_{i = 1}^{N}$, ground truth trajectories $\mathcal{T}_i = \{t_{ij}\}_{j = 1}^{N_i}$, object proposals $\mathcal{D}_i = \{d_{im}\}_{m = 1}^{M_i}$ and their corresponding projections for each multi-video sequence $(v_i^1, \cdots, v_i^K)$}
\BlankLine
\Output{Binary classifiers $\{(\mathbf{w}^k_{Lost}, b^k_{Lost})\}_{k=1}^K$ for data association}
\BlankLine
\Repeat{all targets are successfully tracked}{
\ForEach{multi-video sequence $(v_i^1, \cdots, v_i^K)$ in $\mathcal{V}$}{
\ForEach{target $t_{ij}$}{
Initialize the MDP in an Active state\;
$l \leftarrow$ index of the first frame in which $t_{ij}$ is correctly detected\;
Transfer the MDP to a Tracked state and initialize the target template for each camera view in which target is observed\; 
\While{$l \leq$ index of last frame of $t_{ij}$}{
Fuse object proposals as described in \ref{fusion}\;
Follow the current policy and choose an action $a$\;
Compute the action $a_{gt}$ according to the ground truth\;
\eIf{current state is lost and $a \neq a_{gt}$}{
\ForEach{camera view $k$ in which the target has been seen}{
Decide the label $y^k_{m_k}$ of the pair $(t^k_{ij}, d^k_{im_k})$\;
$\mathcal{S}^k \leftarrow \mathcal{S}^k\ \bigcup\ \{(\phi(t^k_{ij}, d^k_{im_k}), y_{m_k})\}$\;
$(\mathbf{w}^k_{Lost}, b^k_{Lost}) \leftarrow$ solution of Eq.\ref{eq:SVM_lost} on $\mathcal{S}^k$\;
}
break\;
}{
Execute action a\;
$l \leftarrow l + 1$\;
}
\If{$l >$ index of last frame of $t_{ij}$}{
Mark target $t_{ij}$ as successfully tracked\;
}
}
}
}
}
\caption{Reinforcement learning of the binary classifier for data association.}\label{alg:reinforce}
\end{algorithm}\DecMargin{2em}

\paragraph{Policy in Lost States}
In a Lost state, the MDP needs to decide whether to keep the target in a Lost state, or transition it to a Tracked state, or mark it as Inactive. We simply mark a lost target as Inactive and terminate the tracking if the target has been lost for more than $L_{Lost}$ frames. The more challenging task is to make the decision between tracking the target and keeping it as lost. This is treated as a data association problem where, in order to transfer a lost target into a Tracked state, the target needs to be associated with an object proposal, else, the target retains its Lost state. 

\textbf{Data Association:} Let $t$ denote a lost target, and $d$ be an object detection. The goal of data association is to predict the label $y \in \{+1, -1\}$ of the pair $(t, d)$ indicating that the target is linked $(y = +1)$ or not linked $(y = -1)$ to the detection. Assuming that the detection $d$ belongs to camera view $k$, this binary classification is performed using the real-valued linear function $f^k(t, d) = (\mathbf{w}_{Lost}^k)^T \cdot \phi_{Lost}(t, d) + b_{Lost}^k$, where $(\mathbf{w}_{Lost}^k, b_{Lost}^k)$ are the parameters that control the function (for camera view $k$), and $\phi_{Lost}(t, d)$ is the feature vector which captures the similarity between the target and the detection. The decision rule is given by $y = +1$ if $f^k(t, d) \geq 0$, else $y = -1$. Consequently, the reward function for data association in a lost state $s$ given the feature set $\{\phi_{Lost}(t, d_j)\}_{m=1}^M$ is defined as
\begin{equation}
R_{Lost}(s, a) = y(a)\bigg(\vc{\text{max }}{M}{m=1} ((\mathbf{w}_{Lost}^{k_m})^T \cdot \phi_{Lost}(t, d_m) + b_{Lost}^{k_m})\bigg),
\end{equation}
where $y(a) = +1$ if action $a = a_6$, $y(a) = -1$ if $a = a_5$ (see Figure~\ref{fig:MDP}), and $m$ indexes M potential detections for association. Potential detections for association with a target are simply obtained by applying a gating function around the last known location of the target in the global coordinate system. Note that based on which camera view each detection $d_m$ originates from, the appropriate weights $(\mathbf{w}_{Lost}^{k_m}, b_{Lost}^{k_m})$ associated with that view are used. As a result, the task of policy learning in the Lost state reduces to learning the set of parameters $\{(\mathbf{w}_{Lost}^k, b_{Lost}^k)\}_{k=1}^K$ for the decision functions $\{f^k(t, d)\}_{k=1}^K$. 

\begin{table}[t!]
\caption{Features used for data association \cite{xiang2015learning}. We introduce two new features (highlighted in bold) based on the global coordinate positions of targets and detections.}
\centering
\resizebox{0.9\linewidth}{!}{%
\begin{tabular}{ c | c  p{5cm} }
\hline
\textbf{Type} & \textbf{Notation} & \multicolumn{1}{c}{\textbf{Feature Description}} \\ \hline

\multirow{3}{*}{FB error} & \multirow{3}{*}{$\phi_1, \cdots, \phi_5$} & {Mean of the median forward-backward errors from the entire, left half, right half, upper half and lower half of the templates obtained from optical flow}\\ \hline

\multirow{9}{*}{NCC} & \multirow{4}{*}{$\phi_6$} & {Mean of the median Normalized Correlation Coefficients (NCC) between image patches around the matched points in optical flow} \\ \\
 & \multirow{4}{*}{$\phi_7$} & {Mean of the median Normalized Correlation Coefficients (NCC) between image patches around the matched points obtained from optical flow} \\ \hline
 
\multirow{6}{*}{Height ratio} & \multirow{3}{*}{$\phi_8$} & {Mean of ratios of the bounding box height of the detection to that of the predicted bounding boxes obtained from optical flow} \\ \cline{2-3}
 & \multirow{2}{*}{$\phi_9$} & {Ratio of the bounding box height of the target to that of the detection} \\ \hline
 
\multirow{3}{*}{Overlap} & \multirow{3}{*}{$\phi_{10}$} & {Mean of the bounding box overlaps between the detection and the predicted bounding boxes from optical flow}\\ \hline

Score & $\phi_{11}$ & Normalized detection score\\ \hline

\multirow{13}{*}{Distance} & \multirow{4}{*}{$\phi_{12}$} & {Euclidean distance between the centers of the target and the detection after motion prediction of the target with a linear velocity model} \\ \cline{2-3}
 & \multirow{4}{*}{$\boldsymbol{\phi_{13}}$} & \textbf{Lateral offset between last known global coordinate position of the target and that of the detection} \\ \cline{2-3}
 & \multirow{4}{*}{$\boldsymbol{\phi_{14}}$} & \textbf{Longitudinal offset between last known global coordinate position of the target and that of the detection} \\ \hline
 
\end{tabular}
}
\label{tab:lost}
\end{table}

\textbf{Reinforcement Learning:} We train the binary classifiers described above using the reinforcement learning paradigm. Let $\mathcal{V} = \{(v_i^1, \cdots, v_i^K)\}_{i = 1}^{N}$ denote a set of multi-video sequences for training, where $N$ is the number of sequences and $K$ is the total number of camera views. Suppose there are $N_i$ ground truth targets $\mathcal{T}_i = \{t_{ij}\}_{j=1}^{N_i}$ in the $i^{th}$ multi-video sequence $(v_i^1, \cdots, v_i^K)$. Our goal is to train the MDP to successfully track all these targets across all camera views they appear in. We start training with initial weights $(\mathbf{w}^k_0, b^k_0)$ and an empty training set $\mathcal{S}^k_0 = \emptyset$ for the binary classifier corresponding to each camera view $k$. Note that when the weights of the binary classifiers are specified, we have a complete policy for the MDP to follow. So the training algorithm loops over all the multi-video sequences and all the targets, follows the current policy of the MDP to track the targets. The binary classifier or the policy is updated only when the MDP makes a mistake in data association. In this case, the MDP takes a different action than what is indicated by the ground truth trajectory. Suppose the MDP is tracking the $j^{th}$ target $t_{ij}$ in the video $v^k_i$, and on the $l^{th}$ frame of the video, the MDP is in a lost state. Consider the two types of mistakes that can happen: i) The MDP associates the target $t^k_{ij}(l)$ to an object detection $d^k_m$ which disagrees with the ground truth, i.e., the target is incorrectly associated to a detection. Then $\phi(t_{ij}^k(l) , d^k_m)$ is added to the training set $\mathcal{S}^k$ of the binary classifier for camera $k$ as a negative example. ii) The MDP decides to not associate the target to any detection, but the target is visible and correctly detected by a detection $d^k_m$ based on the ground truth, i.e., the MDP missed the correct association. Then $\phi(t_{ij}^k(l) , d_m^k)$ is added to the training set as a positive example. After the training set has been augmented, we update the binary classifier by re-training it on the new training set. Specifically, given the current training set $\mathcal{S}^k = \{(\phi(t^k_m, d^k_m), y^k_m)\}_{m=1}^M$, we solve the following soft-margin optimization problem to obtain a max-margin classifier for data association in camera view $k$:
\begin{multline} \label{eq:SVM_lost}
\vc{\text{min }}{}{\vspace{0.3cm}\mathbf{w}^k_{Lost}, b^k_{Lost}, \mathbf{\xi}} \ \ \frac{1}{2} ||\mathbf{w}^k_{Lost}||^2 + C \sum_{m=1}^M \xi_m \\
\text{s.t. }y^k_m\big((\mathbf{w}^k_{Lost})^T \cdot \phi(t^k_m, d^k_m) + b^k_{Lost}\big) \geq 1 - \xi_m, \xi_m \geq 0, \forall m,
\end{multline}
where $\xi_m, m = 1, \cdots, M$ are the slack variables, and $C$ is a regularization parameter. Once the classifier has been updated, we obtain a new policy which is used in the next iteration of the training process. Note that based on which view the data association is carried out in, the weights of the classifier in that view are updated in each iteration. We keep iterating and updating the policy until all the targets are successfully tracked. Algorithm~\ref{alg:reinforce} summarizes the policy learning algorithm.

\textbf{Feature Representation:} We retain the same feature representation described in \cite{xiang2015learning}, but add two features based on the lateral and longitudinal displacements of the last known target location and the object proposal location in the global coordinate system. This leverages 3D information that is otherwise unavailable in 2D MOT. Table~\ref{tab:lost} summarizes our feature representation.

\IncMargin{1em}
\begin{algorithm}[t!]
\caption{Multi-object tracking with MDPs.}\label{alg:MOT}
\smaller
\SetKwInOut{Input}{input}\SetKwInOut{Output}{output}
\Input{A multi-video sequence $(v^1, \cdots, v^K)$, corresponding object proposals $\mathcal{D} = \{d_m\}_{m = 1}^{M}$ and their projections, learned binary classifier weights $\{(\mathbf{w}_{Active}^k, b_{Active}^k)\}_{k = 1}^K$ and $\{(\mathbf{w}^k_{Lost}, b^k_{Lost})\}_{k=1}^K$}
\Output{Trajectories of targets $\mathcal{T} = \{t_j\}_{j = 1}^{N}$ in the sequence}
\BlankLine
\ForEach{frame $l$ in $(v^1, \cdots, v^K)$}{
Fuse object proposals as described in \ref{fusion}\;
\tcc{process targets in tracked states}
\ForEach{tracked target $t_j$ in $\mathcal{T}$}{
Follow the policy, move the MDP of $t_j$ to the next state\;
}
\tcc{process targets in lost states}
\ForEach{lost target $t_j$ in $\mathcal{T}$}{
\ForEach{proposal $d_m$ not covered by any tracked target}{
Compute $f^{k_m}(t_j, d_m) = (\mathbf{w}^{k_m})_{Lost}^T \cdot \phi(t_j, d_m) + b^{k_m}_{Lost}$\;
}
}
Data association with Hungarian algorithm for the lost targets\;
Initialize target templates for uninitialized camera views using matched (fused) proposals\;
\ForEach{lost target $t_j$ in $\mathcal{T}$}{
Follow the assignment, move the MDP of $t_j$ to the next state\;
}
\tcc{initialize new targets}
\ForEach{proposal $d_m$ not covered by any tracked target in $\mathcal{T}$}{
Initialize a MDP for a new target $t$ with proposal $d_m$\;
\eIf{action $a_1$ is taken following the policy}{
Transfer $t$ to the \textit{tracked} state and initialize the target template for each camera view in which target is observed\;
$\mathcal{T} \leftarrow \mathcal{T}\ \bigcup\ \{t\}$\;
}{
Transfer $t$ to the Inactive state\; 
}
}
}
\end{algorithm}\DecMargin{2em}

\begin{table*}[ht]

\centering
\caption{Quantitative results showing ablative analysis of our proposed tracker.}
\label{tab:results}
\begin{tabular}{@{}ccccccccccc@{}}
\toprule
\multirow{2}{*}{\textbf{Criteria for Comparison}} & \multirow{2}{*}{\textbf{Tracker Variant}} & \multicolumn{2}{c}{\textbf{Sensor Configuration}} & \multicolumn{5}{c}{\textbf{MOT Metrics} \cite{bernardin2008evaluating, milan2013challenges}}\\ 
\cmidrule(lr){3-4}\cmidrule(lr){5-9}
 & & \# of Cameras & Range Sensors & MOTA ($\uparrow$) & MOTP ($\downarrow$) & MT ($\uparrow$) & ML ($\downarrow$) & IDS ($\downarrow$)\\ \midrule\midrule

\multirow{5}{*}{\textbf{\begin{tabular}[c]{@{}c@{}}Number of\\Cameras Used\\(Section \ref{results_cameras})\end{tabular}}}
& - & 2 & \cmark & 73.38 & 0.03 & 71.36\% & 16.13\% & 16\\
& - & 3 & \cmark & 77.26 & 0.03 & 77.34\% & 14.49\% & 38\\
& - & 4 & \cmark & 72.81 & 0.05 & 72.48\% & 20.76\% & 49\\
& - & 4$^{\dagger}$ & \cmark & 74.18 & 0.05 & 74.10\% & 18.18\% & 45\\
& - & 6 & \cmark & 79.06 & 0.04 & 79.66\% & 11.93\% & 51\\
& - & 8 & \cmark & 75.10 & 0.04 & 70.37\% & 14.07\% & 59\\
\midrule

\multirow{2}{*}{\textbf{\begin{tabular}[c]{@{}c@{}}Projection Scheme\\(Section \ref{results_projection})\end{tabular}}}
 & Point cloud based projection & 8 & \cmark & \textbf{75.10} & \textbf{0.04} & \textbf{70.37\%} & \textbf{14.07\%} & \textbf{59}\\
 & IPM projection & 8 & \cmark (for fusion) & 47.45 & 0.39 & 53.70\% & 19.26\% & 152\\
\midrule

\multirow{2}{*}{\textbf{\begin{tabular}[c]{@{}c@{}}Fusion Scheme\\(Section \ref{results_fusion})\end{tabular}}}
 & Point cloud based fusion & 8 & \cmark & \textbf{75.10} & \textbf{0.04} & \textbf{70.37\%} & 14.07\% & \textbf{59}\\
 & Distance based fusion & 8 & \cmark (for projection) & 72.20 & \textbf{0.04} & 68.23\% & \textbf{12.23\%} & 65\\
\midrule

\multirow{2}{*}{\textbf{\begin{tabular}[c]{@{}c@{}}Sensor Modality\\(Sections \ref{results_projection},\ref{results_fusion})\end{tabular}}}
 & Cameras+LiDAR & 8 & \cmark & \textbf{75.10} & \textbf{0.04} & \textbf{70.37\%} & \textbf{14.07\%} & \textbf{59}\\
 & Cameras & 8 & \xmark & 40.98 & 0.40 & 50.00\% & 27.40\% & 171\\
\midrule

\multirow{2}{*}{\textbf{\begin{tabular}[c]{@{}c@{}}Vehicle Detector\\(Section \ref{results_detector})\end{tabular}}}
 & RefineNet\cite{rajaram2016refinenet} & 8 & \cmark & \textbf{75.10} & \textbf{0.04} & \textbf{70.37\%} & \textbf{14.07\%} & \textbf{ 59}\\
 & RetinaNet\cite{lin2017focal} & 8 & \cmark & 73.89 & 0.05 & 68.37\% & 17.07\% & 72\\
 & SubCat\cite{ohn2014fast} & 8 & \cmark & 69.93 & \textbf{0.04} & 66.67\% & 22.22\% & 81\\
\midrule

\multirow{2}{*}{\textbf{\begin{tabular}[c]{@{}c@{}}Global Position\\based Features\\(Section \ref{results_3Dfeatures})\end{tabular}}}
 & with $\{\boldsymbol{\phi_{13}}, \boldsymbol{\phi_{14}}\}$ & 8 & \cmark & \textbf{75.10} & \textbf{0.04} & \textbf{70.37\%} & \textbf{14.07\%} & \textbf{59}\\
 & without $\{\boldsymbol{\phi_{13}}, \boldsymbol{\phi_{14}}\}$ & 8 & \cmark & 71.32 & 0.05 & 64.81\% & 17.78\% & 88\\
  & & & & & & & & \\
\bottomrule
\end{tabular}
\label{tab:results}
\end{table*}

\subsection{Multi-Object Tracking with MDPs} \label{MOT}
After learning the policy/reward of the MDP, we apply it to the multi-object tracking problem. We dedicate one MDP for each target, and the MDP follows the learned policy to track the object. Given a new input video frame, targets in tracked states are processed first to determine whether they should stay as tracked or transfer to lost states. Then we compute pairwise similarity between lost targets and object detections which are not covered by the tracked targets, where non-maximum suppression based on bounding box overlap is employed to suppress covered detections, and the similarity score is computed by the binary classifier for data association. After that, the similarity scores are used in the Hungarian algorithm \cite{munkres1957algorithms} to obtain the assignment between detections and lost targets. According to the assignment, lost targets which are linked to some object detections are transferred to tracked states. Otherwise, they stay as lost. Finally, we initialize a MDP for each object detection which is not covered by any tracked target. Algorithm \ref{alg:MOT} describes our 3D MOT algorithm using MDPs in detail. Note that, tracked targets have higher priority than lost targets in tracking, and detections covered by tracked targets are suppressed to reduce ambiguities in data association.

\section{Experimental Analysis}

\textbf{Testbed:} Since we propose full-surround MOT  using vision sensors, we use a testbed comprising of 8 outside looking RGB cameras (seen in Figure~\ref{fig:motivation}). This setup ensures full surround coverage of the scene around the vehicle, while retaining a sufficient overlap between adjacent camera views. Frames captured from these cameras along with annotated surround vehicles are shown in Figure~\ref{fig:motivation}. In addition to full vision coverage, the testbed has full-surround Radar and LiDAR FoVs. Despite the final goal of this study being full-surround MOT, we additionally consider cases where only a subset of the vision sensors are used to illustrate the modularity of the approach. More details on the sensors, their synchronization and calibration, and the testbed can be found in \cite{rangesh2017multimodal}.

\textbf{Dataset:} To train and test our 3D MOT system, we collect a set of four sequences, each 3-4 minutes long, comprising of multi-camera videos and LiDAR point clouds using our testbed described above. The sequences are chosen much longer than traditional MOT sequences so that long range maneuvers of surround vehicles can be tracked. This is very crucial to autonomous driving. We also annotate all vehicles in the 8 camera videos for each sequence with their bounding box, as well as track IDs. It should be noted that each unique vehicle in the scene is assigned the same ID in all camera views. With these sequences set up, we use one sequence for training our tracker, and reserve the rest for testing. All our results are reported on the entire test set.

\textbf{Evaluation Metrics:} We use multiple metrics to evaluate the multiple object tracking performance as suggested by the MOT Benchmark \cite{bernardin2008evaluating}. Specifically, we use the 3D MOT metrics described in \cite{milan2013challenges}. These include Multiple Object Tracking Accuracy (MOTA), Multiple Object Tracking Precision (MOTP), Mostly Track targets (MT, percentage of ground truth objects who trajectories are covered by the tracking output for at least 80\%), Mostly Lost targets (ML, percentage of ground truth objects who trajectories are covered by the tracking output less than 20\%), and the total number of ID Switches (IDS). In addition to listing the metrics in Table~\ref{tab:results}, we also draw arrows next to each of them  indicating if a high ($\uparrow$) or low ($\downarrow$) value is desirable. Finally, we provide top-down visualizations of the tracking results in a global coordinate system centered on the ego-vehicle for qualitative evaluation.

\begin{figure*}[ht]
  	\centering
	\begin{subfigure}[b]{0.12\textwidth}
		\centering
		\includegraphics[width=\linewidth]{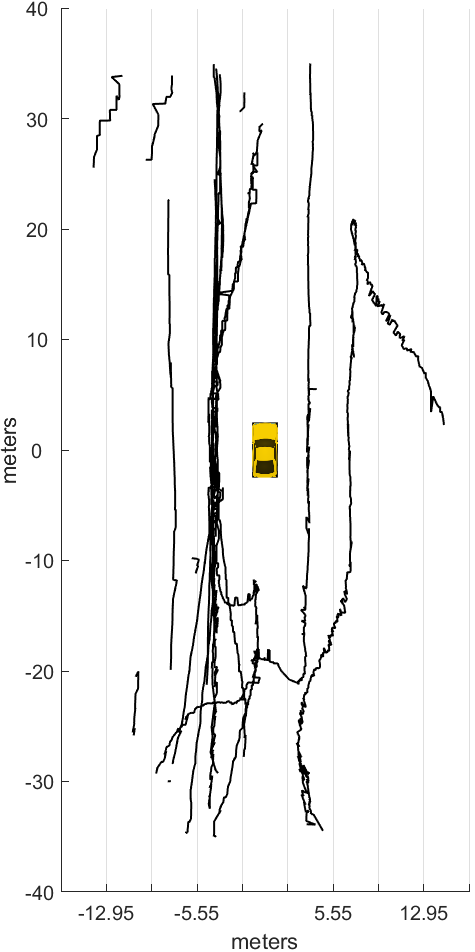}
		\caption{Ground truth}
	\end{subfigure}%
	\hspace{1mm}
~
	\begin{subfigure}[b]{0.12\textwidth}
		\centering
		\includegraphics[width=\linewidth]{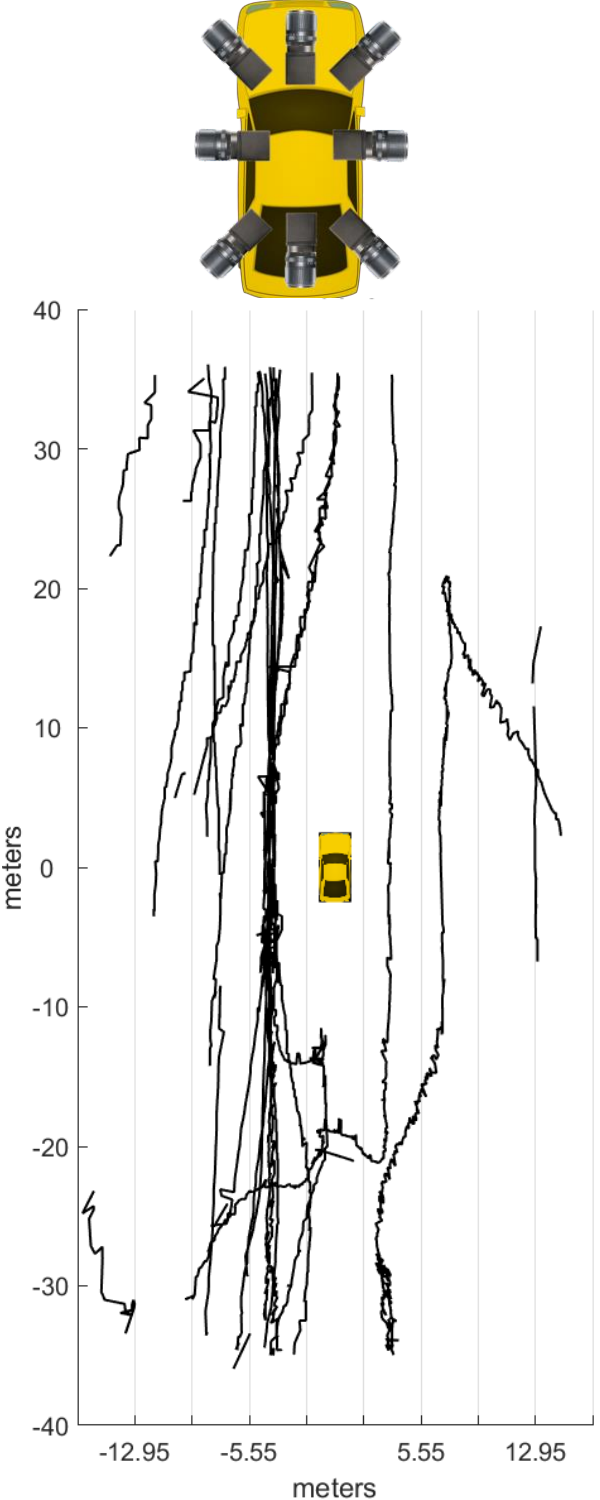}
		\caption{8 cameras}
	\end{subfigure}%
	\hspace{1mm}	
~
	\begin{subfigure}[b]{0.12\textwidth}
		\centering
		\includegraphics[width=\linewidth]{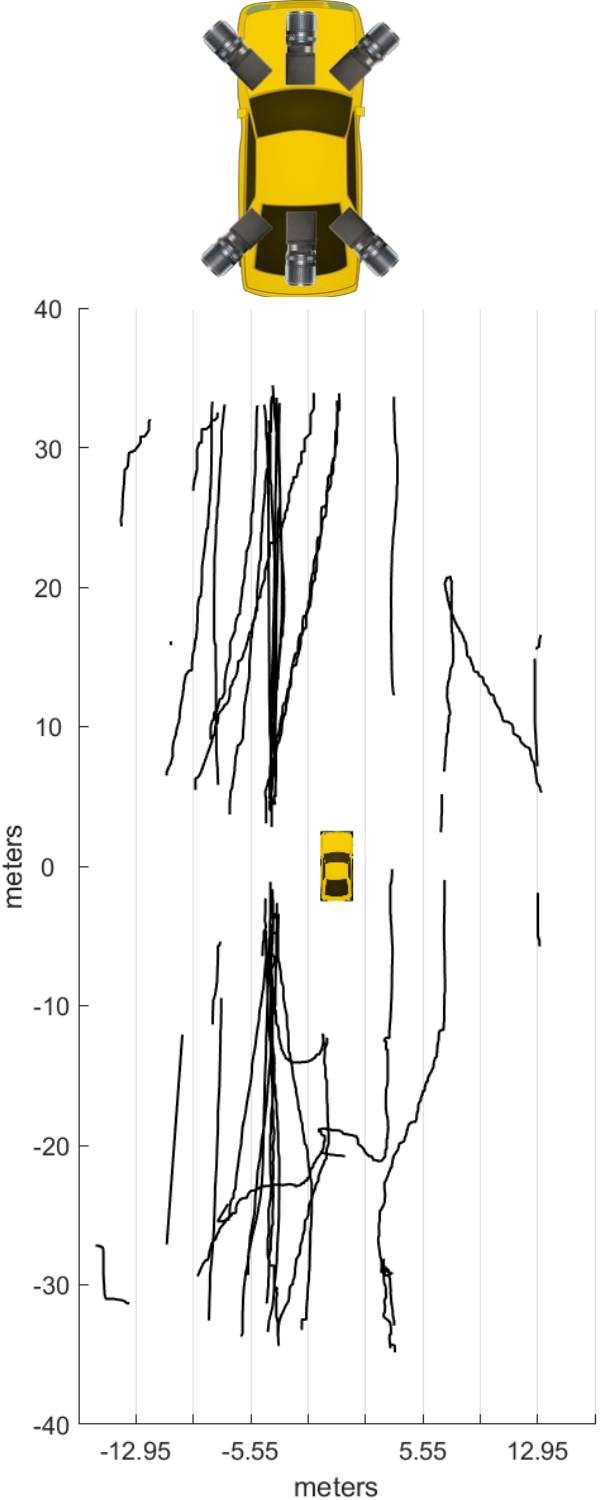}
		\caption{6 cameras}
	\end{subfigure}%
	\hspace{1mm}	
~
	\begin{subfigure}[b]{0.12\textwidth}
		\centering
		\includegraphics[width=\linewidth]{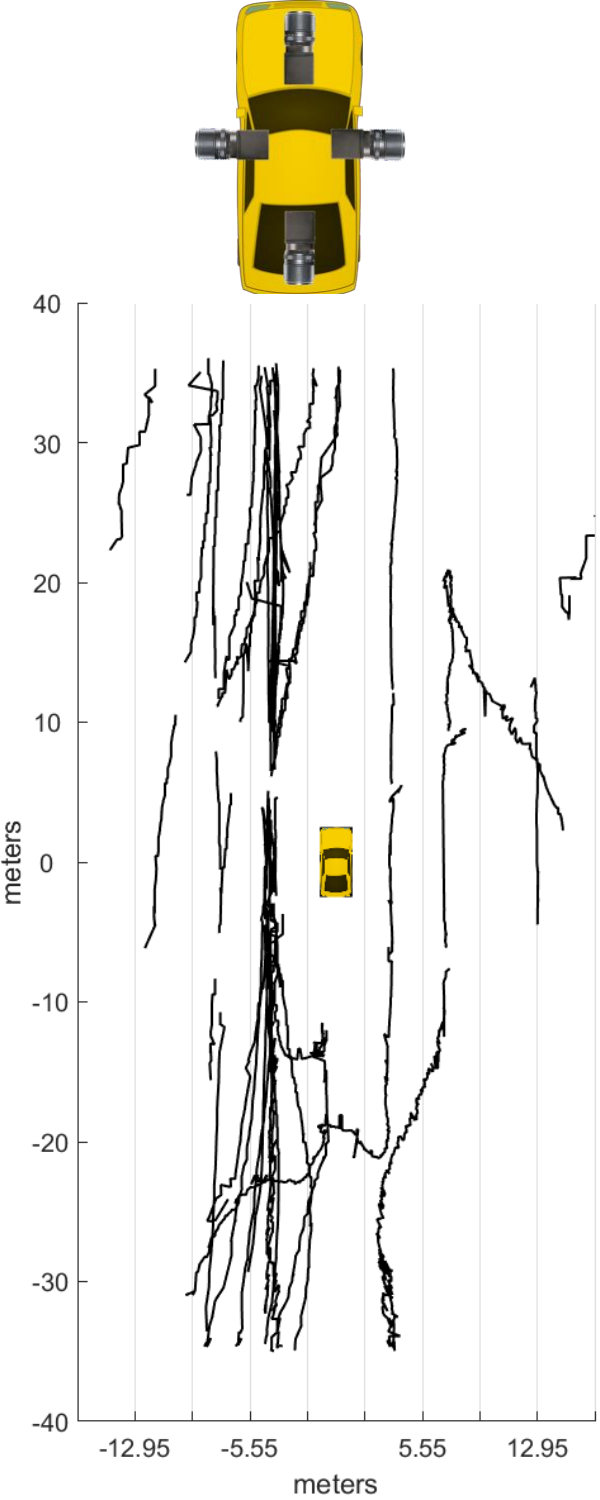}
		\caption{4 cameras}
	\end{subfigure}%
	\hspace{1mm}
~
	\begin{subfigure}[b]{0.12\textwidth}
		\centering
		\includegraphics[width=\linewidth]{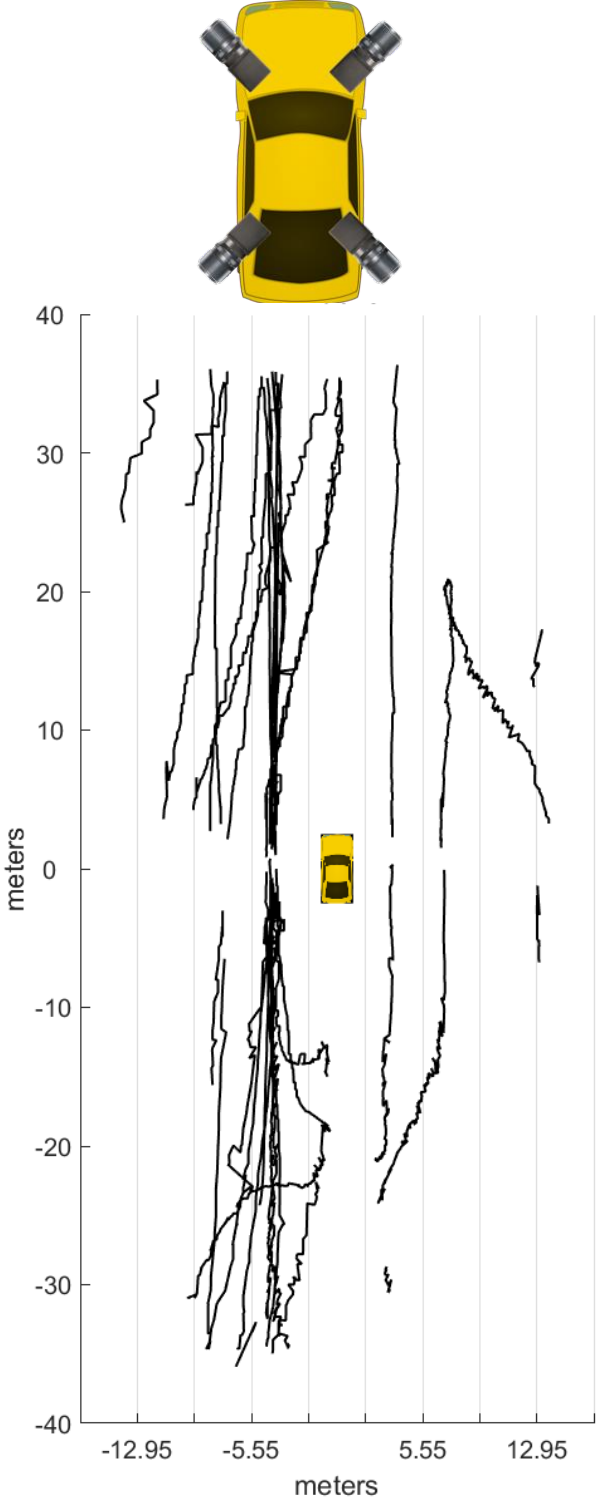}
		\caption{4$^{\dagger}$ cameras}
	\end{subfigure}%
	\hspace{1mm}
~
	\begin{subfigure}[b]{0.12\textwidth}
		\centering
		\includegraphics[width=\linewidth]{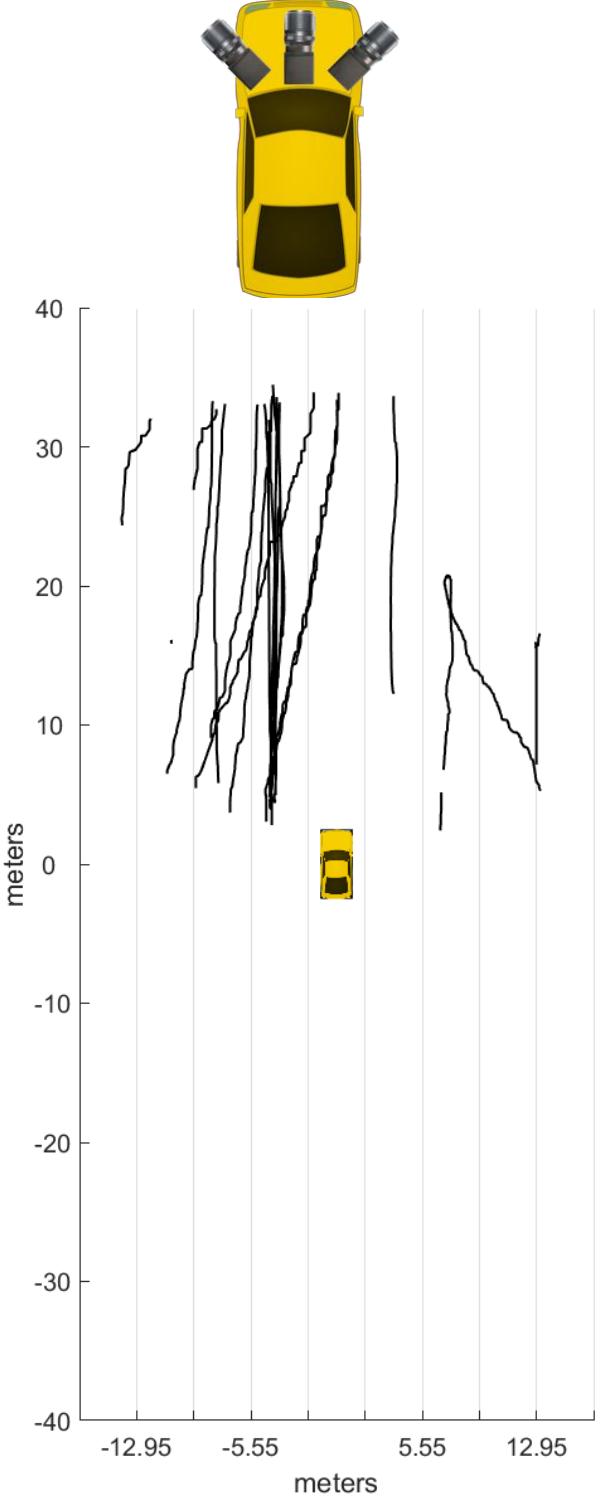}
		\caption{3 cameras}
	\end{subfigure}%
	\hspace{1mm}	
~
	\begin{subfigure}[b]{0.12\textwidth}
		\centering
		\includegraphics[width=\linewidth]{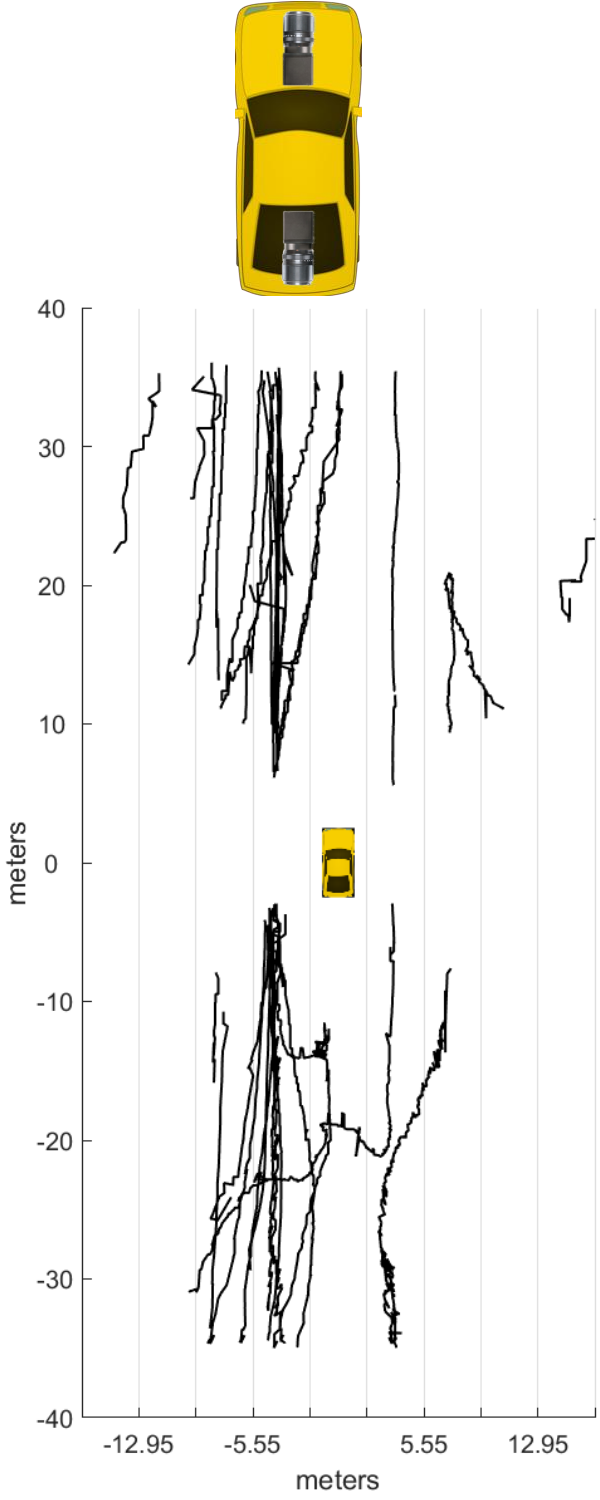}
		\caption{2 cameras}
	\end{subfigure}%
	\hspace{1mm}	
\caption{Tracking results with different number of cameras. The camera configuration used is depicted above each result.}
\label{fig:results_cameras}
\end{figure*}

\subsection{Experimenting with Number of Cameras}\label{results_cameras}

As our approach to tracking is designed to be extremely modular, we test our tracker with different camera configurations. We experiment with 2, 3, 4, 6 and 8 cameras respectively. Top-down visualizations of the generated tracks for a test sequence are depicted in Figure~\ref{fig:results_cameras}. The ground truth tracks are provided for visual comparison. As can be seen, the tracker provides consistent results in its FoV irrespective of the camera configuration used, even if the cameras have no overlap between them.

The quantitative results on the test set for each camera configuration are listed in Table~\ref{tab:results}. It must be noted that the tracker for each configuration is scored only based on the ground truth tracks visible in that camera configuration. The tracker is seen to score very well on each metric, irrespective of the number of cameras used. This illustrates the robustness of the M$^{3}$OT framework. More importantly, it is seen that our tracker performs exceptionally well in the MT and ML metrics, especially in camera configurations with overalapping FoVs. Even though our test sequences are about 3 minutes long in duration, the tracker mostly tracks more than 70\% of the targets, while mostly losing only a few. This demonstrates that our M$^{3}$OT framework is capable of long-term target tracking.

\subsection{Effect of Projection Scheme}\label{results_projection}

\begin{figure}[!h]
\captionsetup[subfigure]{justification=centering}
  	\centering
	\begin{subfigure}[t]{0.12\textwidth}
		\centering
		\includegraphics[width=\linewidth]{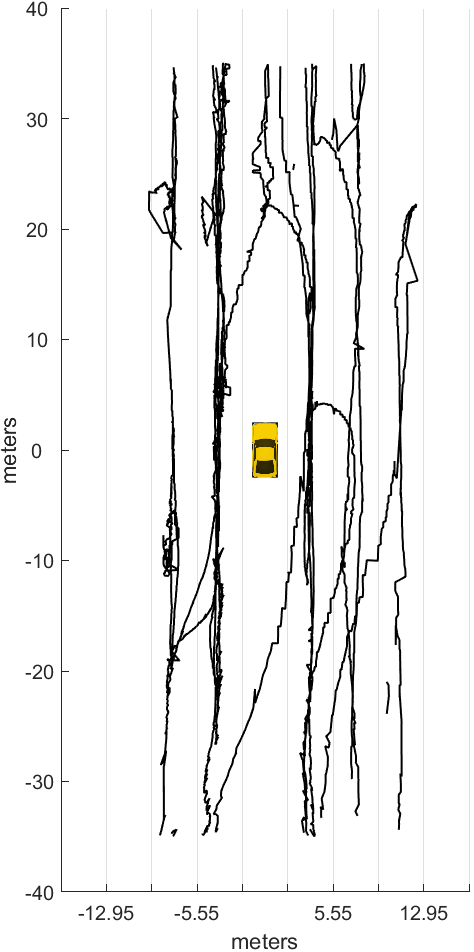}
		\caption{Ground truth}
	\end{subfigure}%
	\hspace{1mm}
~
	\begin{subfigure}[t]{0.12\textwidth}
		\centering
		\includegraphics[width=\linewidth]{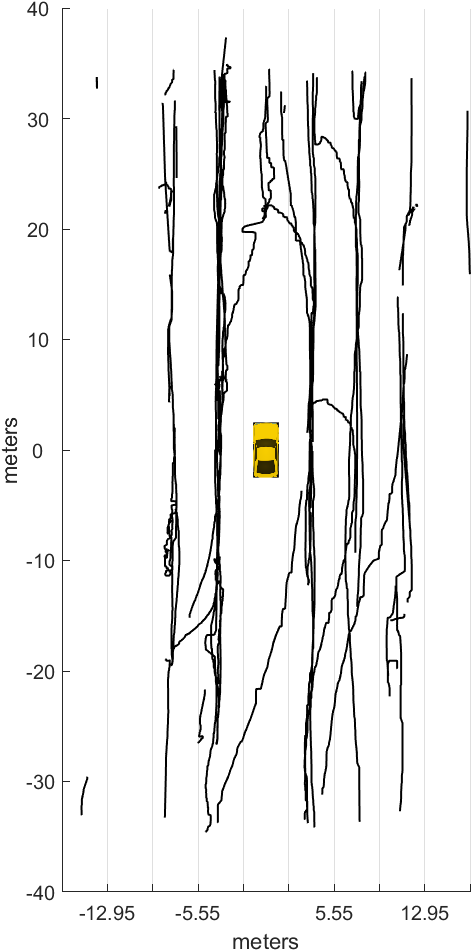}
		\caption{LiDAR based projection}
	\end{subfigure}%
	\hspace{1mm}
~  	
	\begin{subfigure}[t]{0.12\textwidth}
		\centering
		\includegraphics[width=\linewidth]{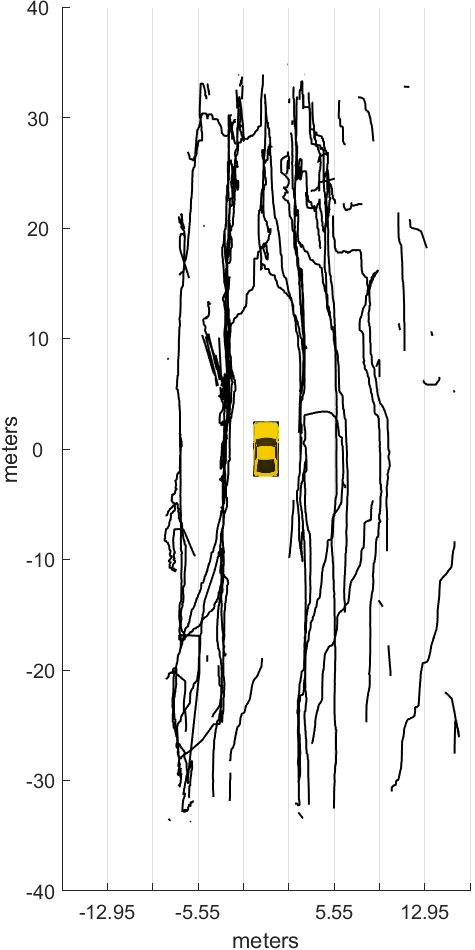}
		\caption{IPM based projection}
	\end{subfigure}
	\caption{Tracking results on a test sequence with different projection schemes.}
	\label{fig:results_projection}
\end{figure}

Figure~\ref{fig:results_projection} depicts the tracking results for a test sequence using the two projection schemes proposed. It is obvious that LiDAR based projection results in much better localization in 3D, which leads to more stable tracks and fewer fragments. The IPM based projection scheme is very sensitive to small changes in the input domain, and this leads to considerable errors during gating and data association. This phenomenon is verified by the high MOTP value obtained for IPM based projection as listed in Table~\ref{tab:results}.

\subsection{Effect of Fusion Scheme}\label{results_fusion}

Once again, we see that the LiDAR point cloud based fusion scheme is more reliable in comparison to the distance based approach, albeit this difference is much less noticeable when proposals are projected using LiDAR point clouds. The LiDAR based fusion scheme results in objects being tracked longer (across camera views), and more accurately. The distance based fusion approach on the other hand fails to associate certain proposals, which results in templates not being stored for new camera views, thereby cutting short the track as soon as the target exits the current view. This superiority is reflected in the quantitative results shown in Table~\ref{tab:results}. The drawbacks of the distance based fusion scheme are exacerbated when using IPM to project proposals, reflected by the large drop in MOTA for a purely vision based system. This drop in performance is to be expected in the absence of LiDAR sensors. However, it must be noted that half the targets are still tracked for most of their lifetime, while only a quarter of the targets are mostly lost.

\begin{figure}[t]
\begin{center}
\includegraphics[width=0.6\linewidth]{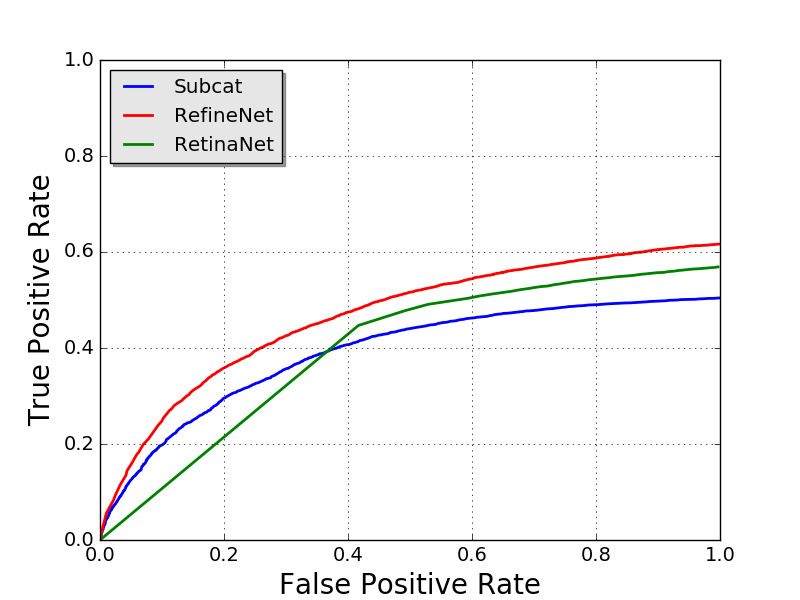}
\end{center}
\caption{ROC curves for different vehicle detectors on the 4 test sequences.}
\label{fig:results_detector}
\end{figure}

\subsection{Effect of using Different Vehicle Detectors}\label{results_detector}

Ideally, a tracking-by-detection approach should be detector agnostic. To observe how the tracking results change for different vehicle detectors, we ran the proposed tracker on vehicle detections obtained from three commonly used object detectors~\cite{rajaram2016refinenet, lin2017focal, ohn2014fast}. All three detectors were trained on the KITTI dataset~\cite{geiger2013vision} and have not seen examples from the proposed multi-camera dataset. The ROC curves for the detectors on the proposed dataset are shown in Figure~\ref{fig:results_detector}. The corresponding tracking results for each detector are listed in Table~\ref{tab:results}. Despite the sub-optimal performance of all three detectors in addition to significant differences in their ROC curves, the tracking results are seen to be relatively unaffected. This indicates that the tracker is less sensitive to errors made by the detector, and consistently manages to correct for it.

\subsection{Effect of Global Position based Features}\label{results_3Dfeatures}
Table~\ref{tab:results} indicates a clear benefit in incorporating features $\{\phi_{13}, \phi_{14}\}$ for data association in Lost states. These features express how near/far a proposal is from the last know location of a target. This helps the tracker disregard proposals that are unreasonably far away from the latest target location. Introduction of these features leads to an improvement in all metrics and therefore justifies their inclusion.

\section{Concluding Remarks}\label{conclusions}
In this work, we have described a full-surround camera and LiDAR based approach to multi-object tracking for autonomous vehicles. To do so, we extend a 2D MOT approach based on the tracking-by-detection framework, and make it capable of tracking objects in the real world. The proposed M$^{3}$OT framework is also made highly modular so that it is capable of working with any camera configuration with varying FoVs, and also with or without LiDAR sensors. An efficient and fast early fusion scheme is adopted to handle object proposals from different sensors within a calibrated camera array. We conduct extensive testing on naturalistic full-surround vision and LiDAR data collected on highways, and illustrate the effects of different camera setups, fusion schemes and 2D-to-3D projection schemes, both qualitatively and quantitatively. Results obtained on the dataset support the modular nature of our framework, as well as its ability to track objects for a long duration. In addition to this, we believe that the M$^{3}$OT framework can be used to test the utility of any camera setup, and make suitable modifications thereof to ensure optimum coverage from vision and range sensors.

\section{Acknowledgments}
We would like to thank the Toyota Collaborative Safety Research Center (CSRC) for their generous and continued support. We would also like to thank our colleagues Kevan Yuen, Nachiket Deo, Ishan Gupta and Borhan Vasli at the Laboratory for Intelligent and Safe Automobiles (LISA), UC San Diego for their useful inputs and help in collecting and labeling the dataset.

\bibliographystyle{IEEEtran}
\bibliography{ref}

%
\begin{IEEEbiography}[{\includegraphics[width=1in,height=1.25in,clip,keepaspectratio]{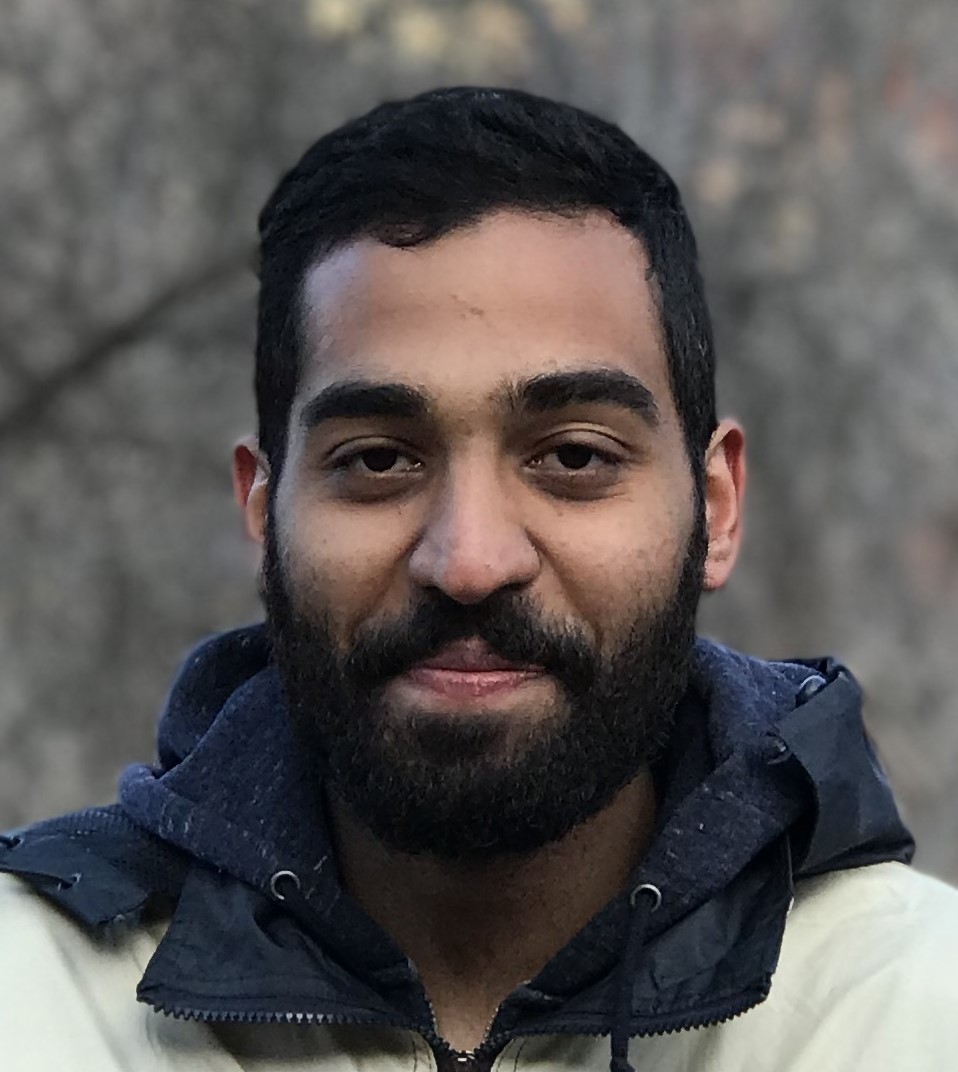}}]{Akshay Rangesh}
is currently working towards his PhD in electrical engineering from the University of California at San Diego (UCSD), with a focus on intelligent systems, robotics, and control. His research interests span computer vision and machine learning, with a focus on object detection and tracking, human activity recognition, and driver safety systems in general. He is also particularly interested in sensor fusion and multi-modal approaches for real time algorithms.
\end{IEEEbiography}

\begin{IEEEbiography}[{\includegraphics[width=1in,height=1.25in,clip,keepaspectratio]{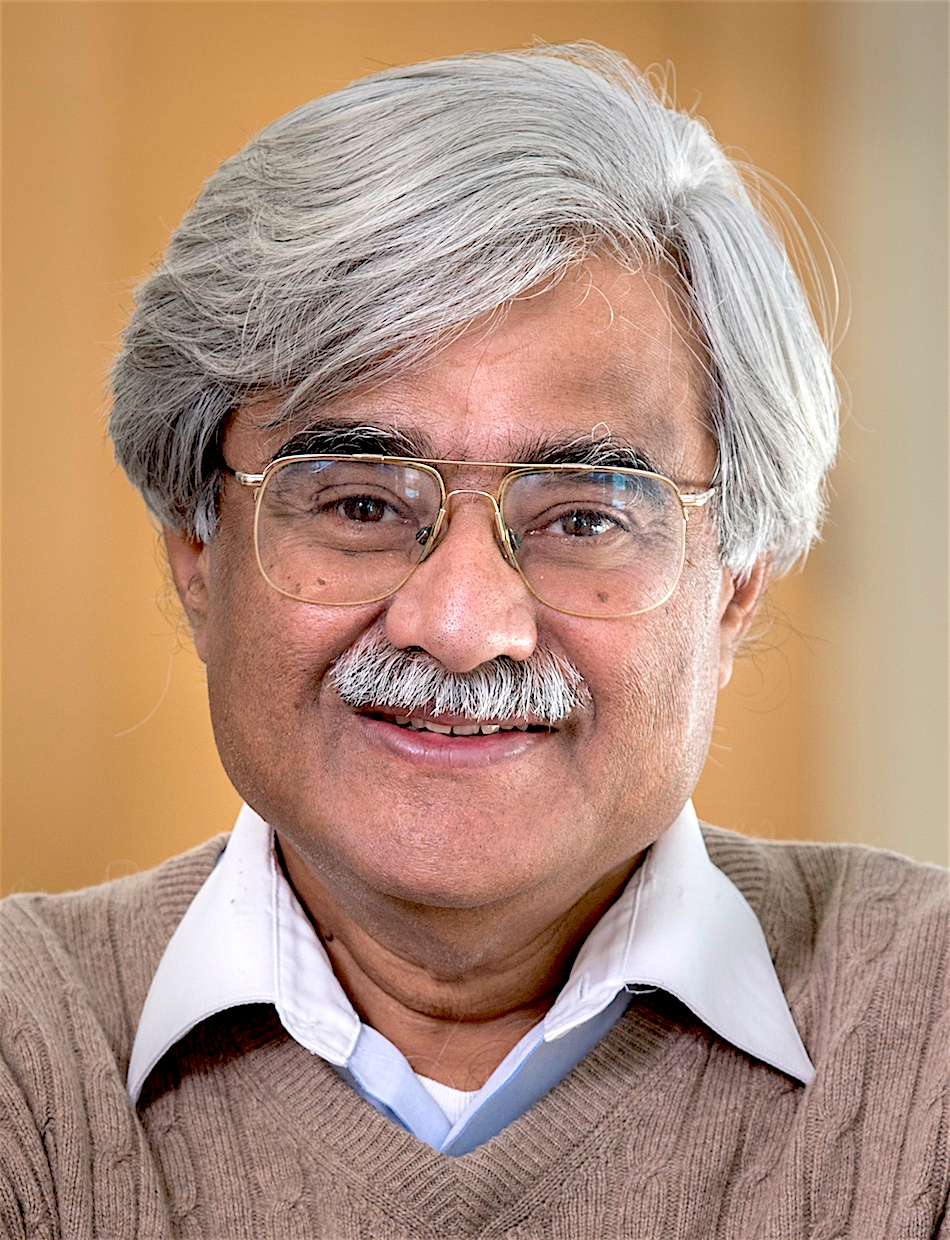}}]{Mohan Manubhai Trivedi}
is a Distinguished Professor at University of California, San Diego (UCSD) and the founding director of the UCSD LISA: Laboratory for Intelligent and Safe Automobiles,
winner of the IEEE ITSS Lead Institution Award (2015). Currently, Trivedi and his team
are pursuing research in intelligent vehicles, machine perception, machine learning, human-robot interactivity, driver assistance, active safety systems. Three of his students have received ``best dissertation" recognitions. Trivedi is a Fellow of IEEE, ICPR and SPIE. He received the IEEE ITS Society's highest accolade ``Outstanding Research Award" in 2013. Trivedi serves frequently as a consultant to industry and government agencies in the USA and abroad. 
\end{IEEEbiography}

\end{document}